%% file: main.tex
\documentclass[lettersize,journal]{IEEEtran}
\usepackage{amsmath,amsfonts}
\usepackage{algorithmic}
\usepackage{algorithm}
\usepackage{array}
\usepackage[caption=false,font=normalsize,labelfont=sf,textfont=sf]{subfig}
\usepackage{textcomp}
\usepackage{stfloats}
\usepackage{url}
\usepackage{verbatim}
\usepackage{graphicx}
\usepackage{cite}
\usepackage{hyperref}
\usepackage{booktabs}
\usepackage{wrapfig}
\usepackage{multicol}
\usepackage{multirow}
\usepackage{makecell}
\usepackage{bbm}
\usepackage{ulem}
\usepackage{color}
\begin{document}

\title{A Comprehensive Survey on Data Augmentation}

\author{
\IEEEauthorblockN{Zaitian Wang}\IEEEauthorrefmark{1},
\IEEEauthorblockN{Pengfei Wang}\IEEEauthorrefmark{1},
\IEEEauthorblockN{Kunpeng Liu},
\IEEEauthorblockN{Pengyang Wang},
\IEEEauthorblockN{Yanjie Fu}, \\
\IEEEauthorblockN{Chang-Tien Lu},
\IEEEauthorblockN{Charu C. Aggarwal},
\IEEEauthorblockN{Jian Pei},
\IEEEauthorblockN{Yuanchun Zhou}\IEEEauthorrefmark{2}

\IEEEcompsocitemizethanks{
    \IEEEcompsocthanksitem \IEEEauthorrefmark{1} Equal contribution. 
    \IEEEauthorrefmark{2} Corresponding author. 
 \IEEEcompsocthanksitem Zaitian Wang, Pengfei Wang, and Yuanchun Zhou are with Computer Network Information Center, CAS and University of Chinese Academy of Sciences. \protect 
 Email: {wangzaitian23@mails.ucas.ac.cn, pfwang@cnic.cn, zyc@cnic.cn}.
 \IEEEcompsocthanksitem Kunpeng Liu is with Department of Computer Science, Portland State University. \protect 
 Email: {kunpeng@pdx.edu}.
 \IEEEcompsocthanksitem Pengyang Wang is with Department of Computer and Information Science, The State Key Laboratory of Internet of Things for Smart City, University of Macau. \protect 
 Email: {pywang@um.edu.mo}.
 \IEEEcompsocthanksitem Yanjie Fu is with Arizona State University. \protect 
 Email: {yanjie.fu@asu.edu}.
 \IEEEcompsocthanksitem Chang-Tien Lu is with Virginia Tech. \protect 
 Email: {ctlu@vt.edu}
 \IEEEcompsocthanksitem Charu C. Aggarwal is with IBM T. J. Watson Research Center. \protect 
 Email: {CharuCAggarwal@gmail.com}
 \IEEEcompsocthanksitem Jian Pei is with Duke University. \protect 
 Email: {j.pei@duke.edu}
 }
        % <-this % stops a space
}

\markboth{Journal of \LaTeX\ Class Files,~Vol.~14, No.~8, August~2021}%
{Shell \MakeLowercase{\textit{Wang et al.}}: A Comprehensive Survey on Data Augmentation}


\maketitle

\input{0_abstract}

\begin{IEEEkeywords}
Data augmentation, Data-centric taxonomy, Multi-modality
\end{IEEEkeywords}

\input{1_introduction}

\input{2_background}
\input{3_taxonomies}
\input{5_1_0_individual}

\input{5_2_0_multiple}

\input{5_3_0_populational}
\input{9_discussion}
\input{10_conclustion}

\section*{Acknowledgments}
This work was supported by the National Natural Science Foundation of China (Grant Nos. 62406306 and 92470204) and the National Key Research and Development Program of China Grant (No. 2024YFF0729201).

% {\appendix[Proof of the Zonklar Equations]
% Use $\backslash${\tt{appendix}} if you have a single appendix:
% Do not use $\backslash${\tt{section}} anymore after $\backslash${\tt{appendix}}, only $\backslash${\tt{section*}}.
% If you have multiple appendixes use $\backslash${\tt{appendices}} then use $\backslash${\tt{section}} to start each appendix.
% You must declare a $\backslash${\tt{section}} before using any $\backslash${\tt{subsection}} or using $\backslash${\tt{label}} ($\backslash${\tt{appendices}} by itself
%  starts a section numbered zero.)}


\bibliography{ref}
\bibliographystyle{IEEEtran}

\newpage

 

% \end{IEEEbiography}



\vfill

\end{document}

%% file: 0_abstract.tex
\begin{abstract}
Data augmentation is a series of techniques that generate high-quality artificial data by manipulating existing data samples. By leveraging data augmentation techniques, AI models can achieve significantly improved applicability in tasks involving scarce or imbalanced datasets, thereby substantially enhancing AI models' generalization capabilities. Existing literature surveys only focus on a certain type of specific modality data and categorize these methods from modality-specific and operation-centric perspectives, which lacks a consistent summary of data augmentation methods across multiple modalities and limits the comprehension of how existing data samples serve the data augmentation process. To bridge this gap, this survey proposes a more enlightening taxonomy that encompasses data augmentation techniques for different common data modalities by investigating how to take advantage of the intrinsic relationship between and within instances. Additionally, it categorizes data augmentation methods across five data modalities through a unified inductive approach. 
\end{abstract}

%% file: 1_introduction.tex
\section{Introduction}

With the rapid development of Artificial Intelligence (AI) in the past decades, AI methods have shown their superiority over human beings and other traditional approaches across most tasks, from universal tasks in our daily lives such as image recognition~\cite{he2016deep} and text translation~\cite{bahdanau2015neural} to complicated scientific research tasks such as drug-drug interaction prediction~\cite{zhang2025motif} and single-cell RNA sequencing analysis~\cite{xu2024sccdcg}. 
Recently, Stable Diffusion~\cite{rombach2022high} and ChatGPT~\cite{openai2023gpt} have greatly changed human beings' working, living, and entertainment. 
The success of AI products is usually attributed to AI models' in-depth understanding of the accumulated data, thereby intrinsically uncovering data patterns and learning data-task correlations.
The performances of AI models will be affected by the quantity and quality of training data, e.g., models trained with limited data suffer from overfitting with significantly degenerative performance on testing datasets, and models trained with imbalanced samples will lead to poor generalization ability. 
In most situations, researchers have to overcome difficulties caused by data scarcity and distribution imbalance. 
An intuitive solution to these problems is to acquire more data, but data are limited or difficult to collect in many cases, and labeling the data is yet another labor-intensive task. 

To solve these problems, data augmentation has been extensively applied and proven to be effective and efficient~\cite{zha2023data}.  
The core idea of data augmentation is to artificially enlarge the training dataset by creating modified copies of existing data. 
It also introduces more diversity and fills the gap between training datasets and real-world applications. 
Data augmentation is widely adopted in the real world and exhibits success in both the dataset preparation stage, such as the ImageNet-C and ImageNet-P datasets~\cite{hendrycks2019benchmarking}, which use data augmentation to simulate common image corruptions and perturbations, as well as the model training stage, such as scSiameseClu~\cite{xu2025scsiameseclu}, which uses data augmentation to generate dual views for graph representation learning. 
As technology continues to evolve, a variety of techniques are applied to augment data samples. 
Some simple approaches only randomly mask part of the data~\cite{devries2017improved,zhong2020random,wei2019eda}, while more sophisticated augmentation schemes involve generative adversarial networks~\cite{bowles2018gan} or reinforcement learning agents~\cite{cubuk2019autoaugment}. 

An ideal survey on data augmentation should be modality-independent, because only in this way can it focus on the intrinsic mechanisms of data augmentation, regardless of the data type, and provide insight into the nature of data augmentation.
Existing literature surveys summarize data augmentation methods from different perspectives~\cite{mikolajczyk2018data,shorten2019survey,khalifa2022comprehensive,mumuni2022data,feng2021survey,shorten2021text,bayer2022survey,ding2022data,zhao2022graph,zhou2022data,borisov2022deep,sauber2022use,fonseca2023tabular,wen2021time,iglesias2023data,iwana2021empirical,cui2024tabular,wang2024survey,sapkota2025image}. As summarized in Table~\ref{tab:surveys}, each survey concentrates on a certain data modality. 
Most of them categorize data augmentation methods with modality-specific (e.g., image modality and text modality) or operation-centric (e.g., feature scaling operation and data perturbation operation) taxonomies. 
These surveys reflect the development of data augmentation and its application in different learning scenarios. 
However, they fail to cover data augmentation methods across data types, uncover their common patterns despite different modalities, and thus limit readers' understanding of the essence of data augmentation. 
For example, originating from the idea of interpolation, mixup-based augmentation has been applied across different modalities, whereas such consistency is ignored in modality-specific discussions. 
To fill this gap, we investigate data augmentation methods comprehensively across the five most popular data modalities. 
We analyze how data augmentation methods utilize different sample numbers and how they leverage different components of the data information. 
Then, we propose a unified taxonomy that focuses on the way data augmentation leverages information from samples. 
Finally, we investigate up-to-date literature and categorize it by our proposed taxonomy.

The main contributions of this survey are as follows: 

\noindent \(\bullet\) We propose a novel modality-independent taxonomy from a data-centric perspective that accommodates data augmentation techniques for all modalities consistently and inductively.

\noindent \(\bullet\) To the best of our knowledge, this is the first survey that covers data augmentation techniques across five modalities: image, text, graph, tabular, and time-series data. 

\noindent \(\bullet\) We investigate how information is consistently contained in each of these modalities and can be utilized for augmentation. 

\noindent \(\bullet\) We include and categorize up-to-date literature in data augmentation and discuss future directions.

\input{tab/intro_surveys}

%% file: tab/intro_surveys.tex
\begin{table}
    \centering
    \caption{A comparison of data modalities and taxonomies discussed by existing surveys and ours. }
    \vspace{-3mm}
    \tiny
    \setlength{\tabcolsep}{1.6pt} % horizontal, Default value: 6pt
    \renewcommand\arraystretch{0}
    \begin{tabular}{cccccccccc}
    \toprule
        & \multicolumn{5}{c}{\textbf{Data Coverage}} & \multicolumn{4}{c}{\textbf{Taxonomies}} \\
        \cmidrule(l){2-6}
        \cmidrule(l){7-10}
        & \quad Image & Text & Graph & Tabular & Time-series \quad & \quad 
        \makecell{Modality-\\specific}
        & \makecell{Modality-\\independent} & \makecell[c]{Operation-\\centric} & \makecell[c]{Data-\\centric} \quad \\ 
        \midrule
        IIPhDW 2018~\cite{mikolajczyk2018data}                 & \checkmark &  &  &  &         & \checkmark & & \checkmark &  \\
        \cmidrule{1-10}
        
        J Big Data 2019~\cite{shorten2019survey}               & \checkmark &  &  &  &                   & \checkmark & & \checkmark &  \\
        \cmidrule{1-10}
         
        IJCNLP 2021~\cite{feng2021survey}          &  & \checkmark &  &  &                   &  & & \checkmark &  \\
        \cmidrule{1-10}
        J Big Data 2021~\cite{shorten2021text}     &  & \checkmark &  &  &                   &  & & \checkmark &  \\
        \cmidrule{1-10}
        IJCAI 2021~\cite{wen2021time}                      &  &  &  &  & \checkmark  & \checkmark & & \checkmark  &  \\
        \cmidrule{1-10}
        Plos one 2021~\cite{iwana2021empirical}            &  &  &  &  & \checkmark & \checkmark & &  \checkmark &  \\

        \cmidrule{1-10}
        Artif Intell Rev 2022~\cite{khalifa2022comprehensive}  & \checkmark &  &  &  &                   &  & & \checkmark &  \\
        \cmidrule{1-10}
        Array 2022~\cite{mumuni2022data}                       & \checkmark &  &  &  &                   & \checkmark & & \checkmark &  \\
        \cmidrule{1-10}
        Comput Surv 2022~\cite{bayer2022survey}    &  & \checkmark &  &  &                   & \checkmark & & \checkmark &  \\
        \cmidrule{1-10}
        KDD 2022~\cite{ding2022data}               &  &  & \checkmark &  &                   & \checkmark & & \checkmark &  \\
        \cmidrule{1-10}
        arXiv 2022~\cite{zhao2022graph}            &  &  & \checkmark  &  &                   &  & & \checkmark & \\
        \cmidrule{1-10}
        arXiv 2022~\cite{zhou2022data}             &  &  & \checkmark &  &                   & \checkmark  & &  \checkmark & \\
        \cmidrule{1-10}
        TNNLS 2022~\cite{borisov2022deep}          &  &  &  & \checkmark &          &  & & \checkmark & \\
        \cmidrule{1-10}
        J Big Data 2022~\cite{sauber2022use}       &  &  &  & \checkmark &          &  & & \checkmark & \\
        \cmidrule{1-10}
        
        J Big Data 2023~\cite{fonseca2023tabular}  &  &  &  & \checkmark &          &  & & \checkmark  &  \\
        \cmidrule{1-10}
        NCA~\cite{iglesias2023data}    &  &  &  &  & \checkmark         &  & & \checkmark & \\
        \cmidrule{1-10}

        arXiv 2024~\cite{cui2024tabular}    &  &  &  & \checkmark &         & \checkmark &  & \checkmark & \\
        \cmidrule{1-10}
        arXiv 2024~\cite{wang2024survey}    &  & \checkmark &  &  &         & \checkmark &  & \checkmark & \\
        \cmidrule{1-10}
        
        arXiv 2025~\cite{sapkota2025image}  & \checkmark & \checkmark &  &  &           &  & \checkmark & \checkmark & \\
        \midrule
        
        \textbf{Ours} & \checkmark & \checkmark & \checkmark & \checkmark & \checkmark &  & \checkmark &  & \checkmark \\

    \bottomrule
    \end{tabular}
    \label{tab:surveys}
    \vspace{-5mm}
\end{table}

%% file: 2_background.tex
\vspace{-4mm}
\section{Background}

\vspace{-2mm}
\subsection{Early Evolution}

Despite the fact that data augmentation only became popular in no more than a decade, some of its embryonic forms were proposed quite early. 
The concept of data augmentation was conceived at the very beginning of deep learning. 
For example, the use of random distortion can be found in LeNet~\cite{lecun1998gradient}, where ninefold the distorted images are added to the dataset to verify that increasing the size of the training set can effectively reduce test error. 
Since then, data augmentation has gradually been recognized as a best practice for training Convolutional Neural Networks (CNNs)~\cite{simard2003best}.
AlexNet~\cite{krizhevsky2012imagenet} explicitly employs data augmentation to reduce overfitting. It augments the datasets by extracting patches from images and altering the color intensity. It also applies dropout, which randomly sets the output of some neurons to zero to prevent their co-adaptation. Although dropout is often regarded as an approach to regularization, it has a similar idea to data augmentation and indeed inspires a set of augmentation methods~\cite{devries2017improved}. 
These applications of data augmentation mainly focus on increasing the size of the training dataset and introducing diversity to it.
SMOTE~\cite{chawla2002smote} is designed from a different perspective. 
Focusing on addressing the class imbalance problem, it suggests that oversampling the minority class can achieve better classification performance when categories are not equally represented in the dataset.

\vspace{-4mm}
\subsection{Existing Surveys}
\vspace{-1mm}

Given the rapid development of CNNs and their wide application in image-related tasks, most early data augmentation methods are proposed for image data.
Also, it is natural to augment image data because many data augmentation methods suit the traits of CNNs very well, such as translation invariance.
Published in 2018, \cite{mikolajczyk2018data} is one of the first surveys on data augmentation. It briefly discusses the effects of traditional and neural-based augmentation methods.
In 2019, data augmentation usage on image data is thoroughly studied in \cite{shorten2019survey}.
It categorizes image data augmentation methods into basic image manipulation and deep learning approaches, with a wide coverage. 
It also introduces how meta-learning selects and combines basic data augmentation operations. 
\cite{khalifa2022comprehensive} is a recent survey on image data augmentation published in 2022. 
It formalizes commonly used operations for image data augmentation with equations and illustrates their effects with examples. 
In the same year, \cite{mumuni2022data} also elaborates on image data augmentation with plenty of examples. 
It adopts taxonomies similar to \cite{shorten2019survey}, but includes more modern approaches and evaluates the improvements in model performance. 

Data augmentation on text data has not been thoroughly researched as early as image data, possibly due to the discrete and correlated nature of text components. 
Still, several augmentation methods are proposed for text data. 
\cite{feng2021survey} is among the first surveys on text data augmentation. It adopts a simple taxonomy but gives more emphasis on the applications. 
Built on previous work on image data augmentation, \cite{shorten2021text} continues to investigate how data augmentation enriches text datasets, and methods are classified as symbolic augmentation and neural augmentation. 
\cite{bayer2022survey} categorizes text data augmentation methods by identifying whether they are used on the data space or feature space. It also presents the performance improvements of different methods. 
\cite{wang2024survey} focuses on the application of data augmentation for training large language models.  
\cite{sapkota2025image} discusses how multimodal large language models can be used for augmenting text, image, and speech data. 

Following the success of data augmentation methods for image and text data, there is a growing trend in augmenting graph data, too. Since the graph represents data in a relatively complex form compared with image and text, there are more opportunities for data augmentation. This also leads to more possible ways to categorize graph data augmentation methods. 
For example, \cite{ding2022data} views graph data augmentation from the employed techniques and application scenarios, while \cite{zhao2022graph} discusses graph data augmentation in terms of its data modality, task level, and whether the augmentation operation is rule-based or learned. 
\cite{zhou2022data} tries to elaborate on all feasible taxonomies for graph data augmentation and uses detailed schematic illustrations to explain the overview framework as well as the idea of some typical methods.

Tabular data is common in reality, but there is not as much room for data augmentation as other data types because tabular data lacks some features for augmentation to utilize. Some early surveys only discuss augmentation techniques in the context of deep learning on tabular data \cite{borisov2022deep}, and others mainly focus on how to generate tabular data \cite{sauber2022use,fonseca2023tabular}. Recently, \cite{cui2024tabular} presented a dedicated and thorough review on tabular data augmentation, focusing on how generative models and retrieval techniques contribute to tabular data augmentation and discussing their future trends. 

With its growing application in scenarios such as speech recognition and IoT, time-series data is attracting more attention. \cite{wen2021time} discusses basic and advanced augmentation methods. \cite{iglesias2023data} focuses on generative approaches. \cite{iwana2021empirical} gives more discussion on mixing time-series data. 

Till now, data augmentation techniques for different data types have been exhaustively discussed; however, there has not been a comprehensive survey that summarizes data augmentation for all types of data. 
Such a survey is profitable because data augmentation techniques for different modalities share some common methodologies when leveraging information from existing data, such as altering some features of a sample or mixing values between multiple samples. Summarizing these similarities can help reveal the common pattern of data augmentation. 
A comprehensive survey does not mean simply combining existing surveys because the proposed taxonomies in these surveys view data augmentation from different angles and are incompatible with each other. 
Some taxonomies contrast basic operations to advanced approaches; some categorize methods with modality-specific terms. 
Besides, these taxonomies provide limited insight into what happens to the information. 
Indeed, huge gaps lie between operations for data modalities due to their distinctive nature; these operation-based or modality-specific taxonomies reflect the development of data augmentation techniques. 
However, from such taxonomies, we cannot grasp the essence of data augmentation and lack perception of the big picture of data augmentation. 
Hence there is a need for a comprehensive survey that covers data augmentation techniques for all data modalities with an all-embracing and data-centric taxonomy. 

%% file: 3_taxonomies.tex
\vspace{-4mm}
\section{Taxonomy}

Before presenting the taxonomy, we formalize data augmentation as follows. Given a labeled dataset $\mathcal{D}_L = \{\mathbf{X},\mathbf{y}\}$, where $\mathbf{X}$ stands for the data, and $\mathbf{y}$ stands for the labels, data augmentation can be represented by a certain function $f_\theta$, such that the augmented dataset $\mathcal{\Tilde{D}}_L = \{\mathbf{\Tilde{X}},\mathbf{\Tilde{y}}\} $ is derived by:
\vspace{-2mm}
\begin{equation}
f_\theta: \mathcal{D}_L = 
    \{\mathbf{X},\mathbf{y}\} 
        \rightarrow 
    \mathcal{\Tilde{D}}_L = 
    \{\mathbf{\Tilde{X}},\mathbf{\Tilde{y}}\} 
\vspace{-2mm}
\end{equation}
This representation also applies to an unlabeled dataset $\mathcal{D}_U = \{\mathbf{X}\}$ or a partially labeled dataset $\mathcal{D}_P = \{\mathbf{X}_L\cup\mathbf{X}_U,\mathbf{y}_L\}$, but for simplicity, we only use a labeled dataset to explain the taxonomy.
Some data augmentation approaches are used in the input data space, while others are used in the latent feature space. This section uses $\mathbf{X}$ as a space-agnostic representation.

In this survey, we try to propose a two-tier taxonomy from a data-centric perspective that can be applied to all data modalities. 
The main consideration of our taxonomy is whence the information in the augmented data comes. 
It differentiates data augmentation techniques by asking two research questions: 
\textbf{RQ1: How many samples are used to generate each new sample? RQ2: Which part of the information is used to generate new data?} 
The answers to these questions constitute the taxonomy hierarchy, which we present below.

By answering \textbf{RQ1}, we can divide data augmentation approaches into \textbf{Single-instance Level Augmentation}, \textbf{Multi-instance Level Augmentation}, and \textbf{Dataset Level Augmentation}, which build the first tier of the taxonomy in our survey. 

\begin{figure*}[htbp]
    \centering
    \includegraphics[width=0.85\linewidth]{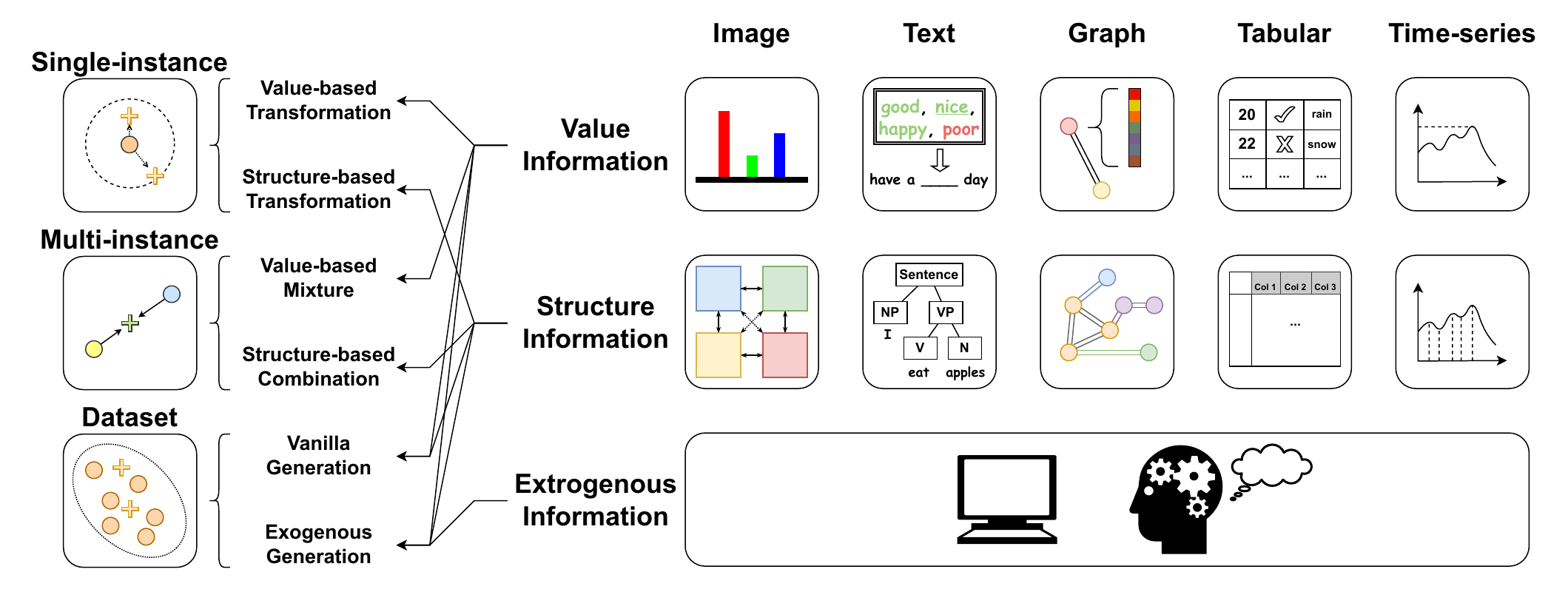}
    \vspace{-5mm}
    \caption{A taxonomy for data augmentation (left) and how information in each data modality is utilized (right).}
    \label{fig:taxonomy}
    \vspace{-5mm}
\end{figure*}

\textit{\textbf{Single-instance Level Augmentation.}}
This type of data augmentation method transforms one data sample without referring to others to create new data. 
It introduces a certain perturbation to the data, such as masking and noise addition. 
The information of new data comes from exactly one sample in the original dataset, and each new sample often lies around its original sample in the feature space. 
We formalize single-instance level augmentation as follows, where $\epsilon(\mathbf{x}_i)$ stands for a certain perturbation related to the original data $\mathbf{x}_i$:
\vspace{-2mm}
\begin{equation}
\mathbf{\Tilde{x}} = \mathbf{x}_i + \epsilon(\mathbf{x}_i),
\qquad
\Tilde{y} = y_i
\vspace{-2mm}
\end{equation}
We further answer \textbf{RQ2} to build the second tier of the taxonomy. 
To answer this question, we must consider how the information is contained in the data.
Typically, data is made up of a certain type of element that carries unit information with a value, such as the coloring of a pixel or the wording in a sentence. 
For most data, there is also some structural information that connects these elements in a certain way, such as the positional relationship of pixels or the syntax of a sentence. 
The key to designing data augmentation methods is to think of how to wisely leverage the two types of information. 
So, we can categorize single-instance level augmentation according to which part of the information is used for data augmentation:
\begin{itemize}
    \item \textbf{Value-based Transformation}, which perturbs the value that an element carries. 
    \item \textbf{Structure-based Transformation}, which perturbs elements' structural relationships. 
\end{itemize}

\textit{\textbf{Multi-instance Level Augmentation.}}
This type of data augmentation method makes use of multiple data to acquire new ones. 
Information from different data samples is combined by operations such as interpolation and concatenation. 
New samples derived from these methods often lie between the source data in the feature space. 
We formalize multi-instance level augmentation as follows, where $i$ and $j$ are two data samples from the original dataset, and $\lambda$ is a scalar factor that often lies between $[0,1]$:
\vspace{-2mm}
\begin{equation}
\mathbf{\Tilde{x}} = \lambda \cdot \mathbf{x}_i + (1-\lambda) \cdot \mathbf{x}_j,
\qquad
\Tilde{y} = \lambda \cdot y_i + (1-\lambda) \cdot y_j
\vspace{-2mm}
\end{equation}
Now we answer \textbf{RQ2} for multi-instance level augmentation. 
Similarly to single-instance level augmentation, this type of data augmentation is also conducted utilizing either value or structure information, and the detailed classification is: 
\begin{itemize}
    \item \textbf{Value-based Mixture}, which arithmetically mixes the values of multiple data. 
    \item \textbf{Structure-based Combination}, which pieces multiple data (or parts of them) together. 
\end{itemize}

\textit{\textbf{Dataset Level Augmentation.}}
This type of data augmentation method does not explicitly make use of one or multiple data points but rather uses the entire dataset.
It comprehensively learns the features and abstractions of the dataset and samples completely new data from a learned distribution of the dataset. 
The generated data should also fall into this distribution. 
We formalize dataset level augmentation as follows, where $P(\cdot)$ stands for a probability distribution learned from the original dataset $\mathcal{D}_L = \{\mathbf{X},\mathbf{y}\}$:
\vspace{-1.5mm}
\begin{equation}
\vspace{-2mm}
\mathbf{\Tilde{x}} \sim P(\mathbf{X}),
\qquad
\Tilde{y} \sim P(\mathbf{y}|\mathbf{\Tilde{x}})
\end{equation}
The answer to \textbf{RQ2} for dataset level augmentation is different from the two categories mentioned above. 
When generating new data based on dataset distributions, most methods simultaneously consider value and structural information. 
Some methods also introduce external knowledge when learning from the source dataset. So, we have the following classification:
\begin{itemize}
    \item \textbf{Vanilla Generation}, where the generation process only resorts to the existing dataset. 
    \item \textbf{Exogenous Generation}, which uses external resources like other datasets or expert knowledge. 
\end{itemize}

We have presented a general taxonomy for data augmentation at a high level with abstract and conceptual formalization. 
We organize this survey in accordance with this taxonomy, from Section \ref{sec:single} to Section \ref{sec:dataset}. 
We elaborate on how the proposed taxonomy applies to each data modality and review some typical data augmentation methods. \footnote{A collection of discussed papers is available at \url{https://github.com/ZaitianWang/Awesome-Data-Augmentation}.}

\begin{table*}[htbp]
\tiny
\centering
\caption{Comparative summary of all methods, including detailed categories, operations, and corresponding analyses. }
\vspace{-2mm}
\setlength{\tabcolsep}{4pt} % horizontal, Default value: 6pt
\renewcommand\arraystretch{0.8} % vertical, Default value: 1
\begin{tabular}{cccclccc}
\toprule
\# of samples & Information type & Modality & Category & Operation & Analysis & Computation & Info loss \\
\midrule
\multirow{36}{*}{Single-instance} & \multirow{14}{*}{Value} & \multirow{2}{*}{Image} & Pixel erasing & Applying a rectangular- or   grid-shaped mask over the image & Moderate & Negligible & High \\
 &  &  & Photometric transformation & Brightness   adjustment, color inversion, color casting, greyscaling, histogram   equalization, kernel filters & Negligible & Moderate & Negligible \\
\cmidrule{3-8}
 &  & \multirow{3}{*}{Text} & Token   replacement & Synonym, embedding similarity,   knowledge graph & Negligible & Moderate & Negligible \\
 &  &  & Token addition & Synonym,   random punctuation & Negligible & Negligible & Negligible \\
 &  &  & Token   deletion & Randomly dropping tokens & Negligible & Negligible & High \\
\cmidrule{3-8}
 &  & Graph & Node attribute masking & Applying   a mask over the node attribute matrix to hide certain nodes or features & Negligible & Negligible & High \\
\cmidrule{3-8}
 &  & \multirow{2}{*}{Tabular} & Table   masking & Applying a mask over the table   to hide some cells & Negligible & Negligible & High \\
 &  &  & Feature engineering & Arithmetic   operations & High & High & Negligible \\
\cmidrule{3-8}
 &  & \multirow{2}{*}{Time-series} & Amplitude perturbation & Scaling, magnitude warping,   rotation, and jittering & Negligible & Negligible & Negligible \\
 &  &  & Sequence decomposition & Extract   components from a sequence & High & Moderate & High \\
\cmidrule{2-8}
 & \multirow{16}{*}{Structure} & \multirow{2}{*}{Image} & Image   cropping & Selecting an area and removing   other parts & Moderate & Negligible & High \\
 &  &  & Geometric   transformation & Flipping,   shearing, rotation, translation, stretching, distortion & Negligible & Moderate & Negligible \\
\cmidrule{3-8}
 &  & \multirow{2}{*}{Text} & Sentence cropping & Dividing sentence to pieces & Moderate & Negligible & High \\
 &  &  & Sentence   morphing & Inversion,   passivization & Moderate & Negligible & Negligible \\
\cmidrule{3-8}
 &  & \multirow{4}{*}{Graph} & Topology perturbation & Node addition, edge addition,   node dropping, edge dropping & Negligible & Negligible & Moderate \\
 &  &  & Subgraphing & Cropping   or substituting subgraphs & Moderate & Negligible & High \\
 &  &  & Graph diffusion & Transition matrices sampled from   random walk & High & Moderate & Moderate \\
 &  &  & Graph   rewiring & Swapping   edges & High & Moderate & Moderate \\
\cmidrule{3-8}
 &  & Tabular & Table subsetting & Splitting table by columns & Negligible & Negligible & High \\
\cmidrule{3-8}
 &  & \multirow{2}{*}{Time-series} & Window   Slicing & Selecting   a span from a sequence & Negligible & Negligible & High \\
 &  &  & Window   Morphing & Downsampling, upsampling,   shuffling & Negligible & Negligible & Negligible \\
\cmidrule{2-8}
 & \multirow{6}{*}{Value-structure} & \multirow{2}{*}{Image} & Policy-based DA & Selecting   an group of augmentation operations with an RL agent & High & High & Moderate \\
 &  &  & Prompt-based image editing & Prompting visual generative   models to edit image & Negligible & High & Negligible \\
\cmidrule{3-8}
 &  & \multirow{3}{*}{Text} & Hierarchical   data augmentation & Augmenting   at both word and sentence level & High & Negligible & Moderate \\
 &  &  & Back-translation & Translating to an intermediate   language and translating back to the original language & Negligible & High & Negligible \\
 &  &  & Paraphrase & Rewriting   sentence with alternative expressions while conveying the same meanings & Negligible & High & Negligible \\
\cmidrule{1-8}
\multirow{24}{*}{Multi-instance} & \multirow{9}{*}{Value} & Image & Image mixup & Mixing color from multiple   images & Negligible & Negligible & Moderate \\
\cmidrule{3-8}
 &  & Text & Text   mixup & Mixing   word or sentence embeddings & Negligible & Moderate & Moderate \\
\cmidrule{3-8}
 &  & Graph & Graph propagation & Spreading node features or   labels through edges to adjcacent nodes & Moderate & Negligible & Moderate \\
\cmidrule{3-8}
 &  & Tabular & Tabular   mixup & Mixing   cell values across rows & Negligible & Moderate & Moderate \\
\cmidrule{3-8}
 &  & Time-series & Sequence averaging & Calculating averaged sequence   values & Negligible & Moderate & Moderate \\
\cmidrule{2-8}
 & \multirow{5}{*}{Structure} & Image & Image Patching & Putting   image segments together & Negligible & Negligible & High \\
\cmidrule{3-8}
 &  & Text & Text fragment merging & Swapping substructures between   sentences & Moderate & Negligible & High \\
\cmidrule{3-8}
 &  & Tabular & Table Concatenation & Unifying   table columns to concatenate tables by overlapping columns & Moderate & Negligible & Negligible \\
\cmidrule{2-8}
 & \multirow{6}{*}{Value-structure} & \multirow{2}{*}{Image} & Neural blending & Mixing images through a neural   network & Negligible & High & Negligible \\
 &  &  & Prompt-based   mixing & Prompting   Diffusion Models to combine images & Negligible & High & Negligible \\
\cmidrule{3-8}
 &  & Text & Generative mixing & Prompting LLMs to mix sentences & Negligible & High & Negligible \\
\cmidrule{3-8}
 &  & Graph & Graph   mixup & Aligning   the topology and mixing node features & Moderate & Moderate & Negligible \\
\cmidrule{1-8}
\multirow{14}{*}{Dataset} & \multirow{4}{*}{Value-structure} & \multirow{4}{*}{Cross-modality} & AE & Using reconstructed sample as   augmentation & Negligible & High & Negligible \\
 &  &  & GAN & Using   the GAN's generator for data augmentation & Negligible & High & Negligible \\
 &  &  & LLM & Prompting the LLM to generate   large amount  of textual samples & Negligible & High & Negligible \\
 &  &  & Diffusion & Using discrete denoising diffusion model  to generate new graphs & Negligible & High & Negligible \\
\cmidrule{2-8}
 & \multirow{9}{*}{Exogenous} & Image, text & Neural Style Transfer & Introducing styles from other   datasets & High & Moderate & Negligible \\
\cmidrule{3-8}
 &  & Tabular & Table   Retrieval & Retrieve   external database to fill and expend existing tables & High & Negligible & Negligible \\
\cmidrule{3-8}
 &  & Image & Computer Graphic Modeling & Designing scenes and objects   with computer softwares & High & High & Negligible \\
\cmidrule{3-8}
 &  & Tabular & Relational   Structure Construction & Building   external graph structures over the table & High & Moderate & Negligible \\
\cmidrule{3-8}
 &  & Time-series & Statistical Generation & Using expert knowledge to expand   sequences & High & Moderate & Negligible \\
 \bottomrule
\end{tabular}
\label{tab:summary}
\vspace{-3mm}
\end{table*}

%% file: 5_1_0_individual.tex
\vspace{-2mm}
\section{Single-instance Level Augmentation}
\label{sec:single}
\vspace{-1mm}

Single-instance level augmentation methods perturb one data sample's value or structure information to obtain the augmented data. 
Different data modalities represent value information with different elemental components and organize these components with different types of structure. 

For image data, the elemental components are pixels. A pixel carries unit information that uses a numerical value to represent the color. It also has spatial relationships with other pixels. Single-instance level image augmentation perturbs the image pixels' colors or spatial relationships.
For text data, the elemental components are words or tokens. A word often expresses minimal meanings. Words are organized into sentences following syntactic rules. 
Single-instance level text augmentation often changes the wording or reshapes the syntax of a sentence.
A graph is represented by $\mathbf{G} = \{\mathcal{V}, \mathcal{E}\}$, where $\mathcal{V}$ and $\mathcal{E}$ stand for the vertex set and the edge set. The topology structure can be alternatively represented by an adjacency matrix $\mathbf{A} \in \mathbb{R}^{|\mathcal{V}|\times|\mathcal{V}|}$. 
The feature matrix $\mathbf{X} \in \mathbb{R}^{|\mathcal{V}| \times d}$ represents the attributes of the nodes. Hence another representation for a graph is $\mathbf{G} = \{\mathbf{A}, \mathbf{X}\}$. Single-instance level graph augmentation perturbs the values in the feature matrix $\mathbf{X}$ or the topology structure represented by the adjacency matrix $\mathbf{A}$.
A table is made up of cells with numerical or categorical values that represent attributes of the described entities. Cells are organized into columns and rows, but this structural information is trivial, and switching the sequence of the columns and rows does not affect the information conveyed by the table. So, most data augmentation methods leverage the value information with only a few exceptions. 
Time-series data represent sequential values within a time span, and each timestamp is considered an elemental component. Data augmentation methods for time-series data either perturb the value at each timestamp or perturb the sequence of timestamps. 

That is, single-instance level augmentation methods focus on utilizing the values and structures of these components.

\vspace{-4mm}
\subsection{Value-based Transformation}
\vspace{-1mm}

\input{5_1_1_individual_value}

\vspace{-3mm}
\subsection{Structure-based Transformation}

\input{5_1_2_individual_structure}

\vspace{-4mm}
\subsection{Value-structure Transformation}

\input{5_1_3_individual_val_struct}

%% file: 5_1_1_individual_value.tex
\subsubsection{Color Perturbation}

\textit{\textbf{Pixel erasing}} masks out all color information in an area of the data; \textit{\textbf{photometric transformation}} imposes a soft adjustment to change the appearance of the image, but the essential information remains visible. Both methods are easy to implement, require none to minimal analysis of the sample, and can be applied to various tasks.

\begin{figure}[htbp]
    \centering
    \hspace*{-2mm}
    \subfloat[]{    
        \includegraphics[width=0.14\linewidth]{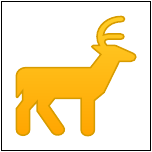}
        \label{subfig:da_img_0}
    } \hspace*{-3mm}
    \subfloat[]{    
        \includegraphics[width=0.14\linewidth]{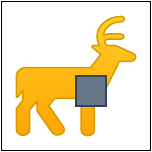}
        \label{subfig:da_img_1}
    } \hspace*{-3mm}
    \subfloat[]{    
        \includegraphics[width=0.14\linewidth]{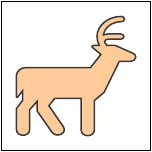}
        \label{subfig:da_img_2}
    } \hspace*{-3mm}
    \subfloat[]{    
        \includegraphics[width=0.14\linewidth]{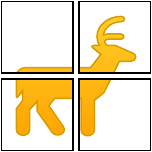}
        \label{subfig:da_img_3}
    } \hspace*{-3mm}
    \subfloat[]{    
        \includegraphics[width=0.14\linewidth]{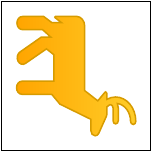}
        \label{subfig:da_img_4}
    } \hspace*{-3mm}
    \subfloat[]{    
        \includegraphics[width=0.14\linewidth]{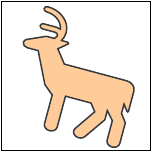}
        \label{subfig:da_img_5}
    } \hspace*{-3mm}
    \subfloat[]{    
        \includegraphics[width=0.14\linewidth]{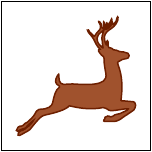}
        \label{subfig:da_img_6}
  }
    \caption{A conceptual demonstration of (a) Original Image, (b) Pixel Erasing, (c) Photometric Transformation, (d) Image Cropping, (e) Geometric
Transformation, (f) Policy-based Data Augmentation (g) Prompt-based Image Editing. }
    \label{fig:image_aug}
\vspace{-2mm}
\end{figure}

\paragraph{\textbf{Pixel Erasing}}
As illustrated in Fig.~\ref{subfig:da_img_1}, pixel erasing masks some information from the original sample. 
Cutout~\cite{devries2017improved} and random erasing~\cite{zhong2020random} select a rectangular region, erase all pixels in that region, and then fill the region with a color or Gaussian noise. 
Hide-and-Seek \cite{kumar2017hide} randomly removes multiple parts of the image so that the model has to learn from multiple parts of the image other than the most discriminating part. 
Grid-Mask~\cite{chen2020gridmask} generates a set of squared masks in the grid layout. 
It can prevent excessive deletion and reservation and achieve a reasonable balance between deleted and reserved regional information.
REF~\cite{hwang2023improving} extends the use of random erasing from the spatial domain to the frequency domain. 
It uses the discrete Fourier transform to process the image, applies random erasing on the image's frequency domain, and uses the inverse discrete Fourier transform to recover the image to the spatial domain. 
Despite its easy application, pixel erasing is subject to removing critical parts from the sample, and some solutions are discussed in Section~\ref{sec:sel_pix}.

\paragraph{\textbf{Photometric Transformation}}
As shown in Fig.~\ref{subfig:da_img_2}, photometric transformation changes the overall color and style of the image.
The simplest way is to adjust the brightness of the image. 
Color inversion replaces the pixels with their ``opposite'' color. 
Color casting applies different operations in different channels. 
Noise injection adds Gaussian or other types of noises to the images. 
Greyscaling transforms a colored picture into black-and-white by combining channels for different colors. 
Histogram equalization~\cite{cheng2004simple} enhances the image by enhancing its contrast, and white balancing simulates alternative light cast on the object. 
Kernel filters are matrices that slide across and are convoluted with the original image to generate new images, either sharpening or blurring the image. 

\vspace{-4mm}

\subsubsection{Wording Perturbation}

\textbf{\textit{Token replacement}}, \textbf{\textit{addition}}, and \textbf{\textit{deletion}} perform word-level manipulation. Some examples are presented in Table~\ref{tab:text_example}. 

\paragraph{\textbf{Token Replacement}}
Token replacement uses alternative wording in a sentence. 
EDA~\cite{wei2019eda} proposes synonym replacement. It randomly selects words from a sentence, substituting them with their synonyms, i.e., words that mean exactly or nearly the same. 
\cite{marivate2020improving} uses WordNet~\cite{miller1995wordnet}, a lexical database, to find synonyms for verbs and nouns and make replacements. 
It also derives word embeddings by Word2Vec~\cite{mikolov2013efficient} and replaces a word with others that have high cosine similarity.
KG-NMT~\cite{moussallem2019augmenting} resorts to a knowledge graph for terminology equivalence to augment samples and improve translation performance.

\paragraph{\textbf{Token Addition}}
Token addition either introduces junk tokens as noise or enriches the sentences to make them more informative. 
The random insertion, proposed in EDA~\cite{wei2019eda}, inserts a random synonym of a random word at a random position in the sentence. 
The words added in the EDA play the role of data noising and may carry either little or harmful information. 
As a variant of EDA, the technique of AEDA~\cite{karimi2021aeda} only inserts punctuation marks. 

\paragraph{\textbf{Token Deletion}}
Random deletion \cite{wei2019eda} randomly removes some words from a sentence. In \cite{xie2017data}, noising techniques are used to avoid overfitting, and the randomly selected word is replaced by a blank placeholder token ``\_", rather than simply being removed. In \cite{duan2023syntax}, apart from word dropout, which is similar to random deletion, it also uses word blanking to replace a word with a special ``<BLANK>" placeholder. 

\subsubsection{\textbf{Node Attributes Masking and Noise Injection}}

Various typical augmentation approaches on node attributes can be applied, from random masking~\cite{you2020graph,feng2020graph,thakoor2021large} to perturbing attributes with random noises~\cite{yang2021graph}. 
These methods are unified by\(\mathbf{\Tilde{X}} = \mathbf{X} \odot \mathbf{M} + \mathbf{N}_W\), where $\mathbf{M} \in \{0,1\}^{N \times N}$ is a binary mask, $\odot$ denotes the Hadamard product, and $\mathbf{N_W} \sim \mathcal{N}(0, \Sigma)$ is the Gaussian white noise.
\cite{song2021topological} first learns the topology information of the graph by random walk and uses the topology embedding together with the original features as inputs to a dual graph neural network. 
LA-GNN \cite{liu2022local} samples features from nodes' neighborhoods as additional information when augmenting the attributes. 
A different approach is proposed in DropMessage \cite{fang2023dropmessage}. Instead of directly modifying the attributes on the nodes, it applies random dropping in the message-passing phase. 

\subsubsection{Cell Value Perturbation}

\textit{\textbf{Table masking}} is a simple technique often applied for tabular data augmentation that corrupts cell values. 
Another approach borrows ideas from \textit{\textbf{feature engineering}}, which automates feature selection and transformation to generate optimal feature space, and the operations can also be used to augment tabular datasets. 

\paragraph{\textbf{Table Masking}}
VIME~\cite{yoon2020vime} uses a mask generator to corrupt features. 
In a self-supervised scenario, two pretext tasks reconstruct the original data and estimate the mask vector, respectively. 
SCARF~\cite{bahri2021scarf} applies data augmentation in a contrastive learning framework. Its corruption process is similar to that of VIME, but it uses InfoNCE loss to evaluate the similarity between the embeddings of the corrupted and original views and then update the model accordingly. 
MET~\cite{majmundar2022met} works in a similar pattern, which calculates the reconstruction loss of a partially masked sample.

\paragraph{\textbf{Feature Engineering}}
Feature transformation and generation techniques can effectively enlarge tabular datasets by generating tables with different feature sets~\cite{he2025fastft,huang2025collaborative,wang2025towards}. 
GRFG \cite{xiao2023traceable} transforms existing features with cascading agents that search for the best arithmetic operation sets. 
MOAT \cite{wang2023reinforcement} extends feature transformation to continuous space by decoding the optimal transformation strategies from continuous embeddings. 
LFG~\cite{zhang2024dynamic} and TIFG~\cite{zhang2024tifg} leverage LLM agents to generate new features that are both explainable and effective. 
TFWT \cite{zhang2024tfwt} computes weights for samples and features, and the weights are combined with the data for later tasks.
Although these studies focus on feature engineering, their way of thinking and application of reinforcement learning are very enlightening for data augmentation tasks.

\subsubsection{Sequence Value Perturbation}

The value information of a sequence is carried on the Y-axis, represented by the amplitude at specific timestamps. \textit{\textbf{Amplitude perturbation}} augments time-series data by altering the Y-axis values, and \textit{\textbf{sequence decomposition}} extracts different parts that express the sequential values from different aspects.

\paragraph{\textbf{Amplitude Perturbation}}
Jittering adds random noise to the value at each timestamp of a sequence.
In \cite{um2017data}, jittering is used to simulate additive noise $\epsilon$ of wearable sensors: 
$\Tilde{x}(\epsilon) = \{x_1+\epsilon_1, \cdots, x_t+\epsilon_t, \cdots, x_T+\epsilon_T\}$.      
Scaling~\cite{um2017data} alters the data by multiplying the magnitude of the entire sequence with a scalar $\alpha$:  
$  \Tilde{x}(\alpha) = \{\alpha x_1, \cdots, \alpha x_t, \cdots, \alpha x_T\}$. 
Magnitude warping~\cite{um2017data} resembles scaling but applies a smooth curve to the sequence so that magnitude at each timestamp is scaled with different parameters: 
$\Tilde{x}(\alpha) = \{\alpha_1 x_1, \cdots, \alpha_t x_t, \cdots, \alpha_T x_T\}$.    
Rotation~\cite{um2017data}, or flipping, assumes the data is symmetric with respect to the X-axis and makes the data upside down by applying a rotation matrix $R$: 
$\Tilde{x}(R) = \{R x_1, \cdots, R x_t, \cdots, R x_T\}$. 
Apart from the amplitude, the phase spectrum provides another option for data augmentation. RobustTAD \cite{gao2020robusttad} converts the time-series data from the time domain to the frequency domain, as shown in equation: $F(\omega_k) 
    = \frac{1}{N} \sum^{N-1}_{t=0}{x_te^{-j\omega_kt}} 
    = A(\omega_k) exp[j\theta(\omega_k)]$, and then increases all phases $\theta(\omega_k)$ by a small perturbation.

\paragraph{\textbf{Decomposition}}
Empirical Mode Decomposition (EMD)~\cite{huang1998empirical} is used to produce a set of intrinsic mode functions. This decomposition relies on extracting the energy linked to different time scales that are inherent to the data. EMD has been utilized as a method of data augmentation in \cite{nam2020data} for the purpose of categorizing impact noise in vehicles.
Seasonal-trend decomposition (STL)~\cite{cleveland1990stl} decomposes signals into seasonality, trend, and remainder:
\(y_t = \tau_t + s_t + r_t, \quad t = 1, 2, \cdots N\)
\cite{wen2019robuststl} and \cite{gao2020robusttad} focus on improving robustness with the help of STL components.

%% file: 5_1_2_individual_structure.tex
\subsubsection{Shape Transformation}

\textit{\textbf{Image cropping}} breaks the original image into pieces (Fig.~\ref{subfig:da_img_3}), and \textit{\textbf{geometric transformation}} perturbs the structure within the original data (Fig.~\ref{subfig:da_img_4}).
Both of them manipulate the spatial relationship between pixels. 

\paragraph{\textbf{Image Cropping}} 
Image cropping selects an area from the original image and removes other parts of the image. 
Attention-based approaches use attention scores to evaluate a subimage. Regions with the main topic often have high attention scores. 
\cite{ma2004automatic} uses the entropy, area ratio, and center distance of the image to evaluate the performance of the image segmentation.
\cite{santella2006gaze} uses eye-tracking techniques to capture human fixations and infer the location of important content on the image. 
Aesthetics-based approaches~\cite{nishiyama2009sensation,cheng2010learning,zhang2012probabilistic} are similar to quality evaluation. 
\cite{nishiyama2009sensation} detects several subject regions and one background region, extracts features from the regions, estimates and combines posterior probabilities, and determines the quality score according to these probabilities. 
A2-RL~\cite{li2018a2} integrates image cropping into a sequential decision-making process and an aesthetics-aware reinforcement learning model to solve the best cropping problem.
Change-based approaches, such as \cite{yan2013learning}, take into account the content that is removed and extract two sets of specially designed features to model what is changed after cropping.

\paragraph{\textbf{Geometric Transformation}}
Geometric transformation changes the spatial relationship of pixels within an image while keeping the value of each pixel untouched. 
Affine transformations like flipping, shearing, rotation, and translation manipulate the image while preserving the colinearity and ratios of distances. 
Image flipping reflects the image around its vertical axis, horizontal axis, or both axes; rotation turns the image by an angle about a fixed point.
Translation means moving an image in a certain direction. It can introduce more variance in the position of target objects in the image. 
Non-affine transformation allows breaking the colinearity and ratio of distances when reshaping the images.
Perspective transformation~\cite{wang2019perspective} mimics images taken from other views of the object. 
Image stretching~\cite{wang2008optimized}  preserves or even enlarges the prominent contents and compresses the background or unimportant foreground information. 
Distortion~\cite{simard2003best,jaderberg2015spatial} focuses on stretching the pixels in a certain region. 
PatchShuffle~\cite{kang2017patchshuffle} divides an image into block matrices and shuffles the pixels within each block under a certain probability. 

\subsubsection{Syntax Transformation}

We categorize syntax transformation as: \textit{\textbf{sentence cropping}} that splits a sentence into many pieces and \textit{\textbf{sentence morphing}} that swaps the location of words or grammatical components. Some examples are presented in Table~\ref{tab:text_example}. 

\paragraph{\textbf{Sentence Cropping}}
Sentence cropping \cite{csahin2018data} divides sentences into small pieces. 
It uses dependency trees to extract the relationships between subjects and objects of the sentence and produces multiple smaller sentences that focus on different parts of the original sentence. 
This operation breaks the syntactic solidity and semantic comprehensiveness of the entire sentence but preserves local syntactic tags and shallow semantic labels. 

\paragraph{\textbf{Sentence Morphing}}
These methods manipulate sentence structure by altering the position of words or terms. 
Random swap~\cite{wei2019eda} randomly chooses two words in a sentence and swaps their position. 
\cite{min2020syntactic} studies two types of syntactic augmentations: inversion and passivization. 
Inversion swaps the subject and object in a sentence, and the semantics might be different in most cases. 
Passivization converts a sentence to its passive form, and the meaning of the new sentence is typically consistent with the original one. 
Sentence rotation~\cite{csahin2018data} chooses the root of the sentence's dependency tree as the center (as in an image) and swaps other fragments.

\begin{table*}
    \centering
    \tiny
    \vspace{-3mm}
    \caption{Demonstration of some elementary operations for text data augmentation}
    \vspace{-3mm}
    \renewcommand{\arraystretch}{0}
    
    \setlength{\tabcolsep}{3pt} % horizontal, Default value: 6pt
    \begin{tabular}{p{3cm}p{13cm}}
        \toprule
        \textbf{Operation} & \textbf{Sentence} \\
        \midrule
        Original & I am an art student and I paint a lot of pictures. Many people pretend that they understand modern art. \\
        \cmidrule{1-2}
        Synonym Replacement \cite{wei2019eda} & I am an art student and I \textbf{\textit{draw}} a lot of pictures. Many people pretend that they understand \textbf{\textit{contemporary}} art. \\
        \cmidrule{1-2}
        Similarity Replacement \cite{marivate2020improving} & I am an art student and I paint a lot of \textbf{\textit{photos}}. Many people pretend that they understand modern \textbf{\textit{literature}}. \\
        \cmidrule{1-2}
        Random Insertion \cite{wei2019eda} & I am an art student and \textbf{\textit{picture}} I paint a lot of pictures. Many \textbf{\textit{appreciate}} people pretend that they understand modern art. \\
        \cmidrule{1-2}
        Punctuation Insertion \cite{karimi2021aeda} & I am an\textbf{\textit{,}} art student and\textbf{\textit{:}} I paint a lot of pictures. Many people\textbf{\textit{?}} pretend that they\textbf{\textit{.}} understand modern art. \\
        \cmidrule{1-2}
        Random Swap \cite{wei2019eda} & I am an \textbf{\textit{of}} student and I paint a lot \textbf{\textit{art}} pictures. Many people \textbf{\textit{they}} that \textbf{\textit{pretend}} understand modern art. \\
        \cmidrule{1-2}
    Random Deletion \cite{wei2019eda} & I am an art \xout{student} and I paint a lot of pictures. Many \xout{people} pretend that they understand modern art. \\
        \cmidrule{1-2}
        Placeholder Replacement \cite{xie2017data} & I am an art \textbf{\textit{\_}} and I paint a lot of pictures. Many \textbf{\textit{\_}} pretend that they understand modern art. \\
        \cmidrule{1-2}
        Inversion \cite{min2020syntactic} & I am an art student and \textbf{\textit{a lot of pictures paint me}}. Many people pretend that they understand modern art. \\
        \cmidrule{1-2}
        Passivization \cite{min2020syntactic} & I am an art student and I paint a lot of pictures. Many people pretend that \textbf{\textit{modern art are understood by them}}. \\
        \bottomrule
    \end{tabular}
    \label{tab:text_example}
    \vspace{-3mm}
\end{table*}

\subsubsection{Topology Transformation}

Graph topology structure has immense information representation capacity. 
\textit{\textbf{Naive topology perturbation}} methods modify a node or an edge. 
\textit{\textbf{Subgraphing}} extracts subgraphs from the complete graph.
\textit{\textbf{Graph diffusion}} and \textit{\textbf{graph rewiring}} are more sophisticated techniques that focus on adjusting the edges to obtain a better graph. 

\paragraph{\textbf{Naive Topology Perturbation}}
Due to their straightforward idea and easy implementation, \textit{\textbf{node dropping}} and \textit{\textbf{node addition}} are among the earliest augmentation approaches in graph learning. 
Node dropping~\cite{you2020graph} randomly discards some nodes from the graph: \(
    \Tilde{\mathcal{V}} = \text{Drop} (\mathcal{V}),
    \Tilde{\mathbf{G}} = \{\Tilde{\mathbf{A}},\Tilde{\mathbf{X}}\}
        = \{\mathbf{A}[\Tilde{\mathcal{V}},\Tilde{\mathcal{V}}], \mathbf{X}[\Tilde{\mathcal{V}}, :]\}
\).
As for node addition, an example is MPNN \cite{gilmer2017neural}. It introduces a latent master node that is connected to all other nodes, which helps messages pass through the graph network.
An edge describes the relationship between two nodes, and by perturbing these relationships through \textit{\textbf{edge dropping}} and \textit{\textbf{edge addition}}, we also introduce diversity to the structural information of a graph.
DropEdge~\cite{rong2019dropedge} randomly sets a portion of the adjacency matrix to zero. 
Edge masking in \cite{thakoor2021large} constructs a Bernoulli distribution and randomly masks edges under that probability. 
FairDrop~\cite{spinelli2021fairdrop} computes a mask according to the network's homophily and uses this mask to guide the dropping process to remove unfair connections. 
AD-GCL~\cite{suresh2021adversarial} applies a GNN augmenter to learn how to drop edges so that the mutual information of the augmented graph and that of the original graph is minimized. 
Despite its popularity, graph topology perturbation is potentially risky in some cases and must be used carefully. The removal of some crucial edges might influence the connectivity of the graph, thereby limiting the use of some graph algorithms that require the graph to be fully connected. Conversely, adding some inter-class edges might escalate representation collapse caused by the inductive bias of the GNN models. 

\paragraph{\textbf{Subgraphing}}
\textit{\textbf{Subgraph cropping}} assumes graph semantics are partially represented in local structures~\cite{you2020graph}. 
GCC~\cite{qiu2020gcc} and SUGAR~\cite{sun2021sugar} use random walk to sample a subgraph from the original graph. 
GraphCrop~\cite{wang2020graphcrop} applies a node-centric strategy to crop connectivity-preserved subgraphs and expand the dataset with these augmented subgraphs. 
SUBG-CON~\cite{jiao2020sub} uses the Personalized PageRank algorithm to select the important nodes. 
NeuralSparse~\cite{zheng2020robust} proposes to obtain the subgraph by learning. It uses a sparsification network to generate sparsified subgraphs. 
\textit{\textbf{Subgraph Substitution}} perturbs the substructure as a whole of a graph.
It is especially beneficial for molecular graphs, which represent atoms with nodes and chemical bonds with edges. 
Functional groups, often deciding the chemical characteristics of molecules, can be represented by substructures. 
MoCL~\cite{sun2021mocl} utilizes substructure information as one of its augmentation schemes. 
Its replacement of functional groups effectively introduces reasonable noise. 
GEAR~\cite{liu2022graph} also employs such a technique. 
It augments molecule graphs by environment replacement, introducing noise from environment subgraphs. 

\paragraph{\textbf{Graph Diffusion}}
Traditional message passing in graph neural networks only leverages one-hop connections, while the edges of a real graph are often noisy or arbitrarily defined, and many relationships in the networks are indirectly contained in higher-order neighborhoods. To address this issue, GDC~\cite{gasteiger2019diffusion} proposes to calculate transition matrices $\mathbf{T}^k$ by random walk and combines these matrices with weights $\theta_k$ to obtain a new adjacency matrix $\mathbf{\Tilde{A}}$, as in:\(\mathbf{\Tilde{A}} = \sum_{k=0}^{\infty} \theta_k\mathbf{T}^k\).
The new graph can also be understood as being processed by a denoising filter because graph diffusion smooths out the neighborhood over the graph~\cite{gasteiger2019diffusion}. 
This technique is widely used by self-supervised approaches to transform the graph and generate different views for contrastive learning, such as MVGRL~\cite{hassani2020contrastive}, SelfGNN~\cite{kefato2021self}, and MERIT~\cite{jin2021multi}.

\paragraph{\textbf{Graph Rewiring}} 
Graph rewiring is initially proposed as randomly swapping edges~\cite{schneider2011mitigation} to improve a network's robustness and mitigate malicious attacks, as formulated by:\(    \mathbf{\Tilde{A}} = 
        \mathbf{A} \odot (\mathbf{1} - \mathbbm{1}_r) + 
        (\mathbf{1} - \mathbf{A}) \odot \mathbbm{1}_r\). 
Smart rewiring~\cite{louzada2013smart} takes the degrees of chosen nodes when swapping their edges. 
Stochastic Discrete Ricci Flow~\cite{topping2021understanding} iteratively adds edges that can improve the most negatively curved edge and then removes the edge with the most positive curvature. 
DiffWire~\cite{arnaiz2022diffwire} learns the commute times and spectral gap of the graph. 
It rewires the adjacency graph so that the graph's spectral gap is minimized. 
DHGR~\cite{bi2022make} rewires the graph by adding edges between high-similarity node pairs and removing edges from low-similarity ones. 
It can effectively improve the homophily of the graph and thus improve GNN performance. 
HDHGR~\cite{guo2023homophily} computes multiple similarity matrices under different meta-paths and thus generates several rewired meta-path subgraphs. 

\subsubsection{\textbf{Table Subsetting}}
Inspired by image cropping~\cite{chen2016automatic,yan2013learning}, SubTab~\cite{ucar2021subtab} divides a table into multiple subsets by splitting its columns. Each subset contains only a percentage of all features. The model is forced to learn the representation of the samples by these features and is required to reconstruct all the features from the subset. 

\subsubsection{Sequence Transformation}

Unlike value perturbation, sequence perturbation is implemented on the X-axis.
\textit{\textbf{Window slicing}} splits the time-series data into pieces. 
It is also the basis of many \textit{\textbf{sequence morphing}} methods because they apply varied operations on these slices. 

\paragraph{\textbf{Window Slicing}}
Inspired by image cropping, MCNN~\cite{cui2016multi} proposes window slicing. It extracts slices from the entire sequence as new training samples. Each slice is assigned the label of the original sequence. 

\paragraph{\textbf{Sequence Morphing}}
Based on window slicing, \cite{le2016data} extends or compresses the generated spans, which is called window wrapping. This is equivalent to speeding up or slowing down a span of the sequence.
Vowel stretching~\cite{nagano2019data} is a special kind of window wrapping technique. It only perturbs the sequence by prolonging the vowels. This is meant to be consistent with how children speak.
Random shuffling~\cite{um2017data,steven2018feature}, randomly permutates these slices and concatenates them as a long sequence.
VTLP~\cite{jaitly2013vocal} applies window wrapping in the frequency domain. It adds a small distortion to the central frequency. 
SpecAugment~\cite{park2019specaugment} also augments the frequency domain. It masks consecutive frequency channels of length sampled from a uniform distribution.

%% file: 5_1_3_individual_val_struct.tex
\subsubsection{Image Value-structure Transformation}
\textit{\textbf{Policy-based}} methods search for the optimized augmentation policy that includes both value-based and structure-based transformations, as demonstrated in Fig.~\ref{subfig:da_img_5}. \textit{\textbf{Prompt-based image editing}} employs a text-to-image model to change the content, background, or style of the image as humans desire, changing both the coloring and the position of pixels (Fig.~\ref{subfig:da_img_6}). 

\paragraph{\textbf{Policy-based Data Augmentation}}
A policy includes a sequential group of basic augmentation operations with probabilities and magnitudes. 
AutoAugment~\cite{cubuk2019autoaugment} represents data augmentation as a decision-making process and uses reinforcement learning to search for the best policy. 
It processes images under the policy, evaluates these augmented images on downstream networks, and rewards a policy generator for better performance.
Fast AutoAugment~\cite{lim2019fast} proposes a more efficient searching algorithm, which is faster and thus allows a larger number of sub-policies, benefiting the generalization performance. 
These methods rely on independent reinforcement learning processes for each task, and the searched policies do not guarantee optimality when transferred to other tasks, making the methods cost-inefficient. 
RandAugment~\cite{cubuk2020randaugment} removes the need for a separate search phase on a proxy task and makes the training process less complicated and less expensive. 
Population Based Augmentation~\cite{ho2019population} learns an augmentation schedule rather than a fixed policy as in AutoAugment to improve efficiency. 
These methods combine the benefits of both value-based transformations and structure-based transformations and further discover the potential of data augmentation with individual image data. 

\paragraph{\textbf{Prompt-based Image Editing}}
Visual generative models are typically conditioned on text embeddings, allowing flexible image editing by modifying textual descriptions of images and prompting the models. 
DA-Fusion~\cite{trabucco2024effective} learns fine-grained concepts, replaces the original category labels, and uses the Stable Diffusion model to generate new images. 
SaSPA~\cite{michaeli2024advancing} uses GPT-4 to generate prompts based on the sample's meta-class, then inputs the prompts, reference images, and detected edges to the diffusion model to generate augmented images. 
Generating the final augmentation set with diffusion models can be labor-intensive because models are very sensitive to the prompts, so users must prepare detailed image descriptions carefully and might have to examine the prompts repeatedly. 

\subsubsection{Text Value-structure Manipulation}
\textit{\textbf{Hierarchical data augmentation}} explicitly augments both value and structure information of text data. \textit{\textbf{Back-translation}} and \textit{\textbf{paraphrase augmentation}} use neural networks to generate sentences that are semantic-preserving but different from original sentences in both wording and syntax. 

\paragraph{\textbf{Hierarchical Data Augmentation}}
% \textit{\textbf{Hierarchical Data Augmentation.}}
Hierarchical data augmentation~\cite{yu2019hierarchical} augments text data at both word and sentence levels. 
It evaluates the roles of words and sentences by training a hierarchical attention network to obtain the attention values of the text data at both levels. 
Attention values help decide the importance of components and augment the data by cropping and concatenating these components. 

\paragraph{\textbf{Back-translation}}
\cite{sennrich2016improving} proposes to translate text data from one language to another, or a sequence of other languages, and finally translates back to the original language. 
Semantic information is preserved, but the translation process can introduce some variance in syntax and wording. 
Such transformations can help the model learn more flexible expressions that convey the same meaning. 

\paragraph{\textbf{Paraphrase Augmentation}}
AugGPT~\cite{dai2025auggpt}, for example, prompts the state-of-the-art large language model (LLM) to rewrite sentences in the dataset and preserves dataset coherence after data augmentation. 
Taboo~\cite{cegin2024effects} requires paraphrasing without using some frequently used words, Chain~\cite{cegin2024effects} paraphrases one sentence multiple times, and Hint~\cite{cegin2024effects} provides some examples to encourage the LLM response otherwise. DoAug~\cite{wang2025diversity} constructs a preference dataset based on the diversity of paraphrased sentences and trains the LLM with the DPO to boost its generation diversity. 
Although such paraphrase methodology is capable of preserving sentence meanings, it is still possible that the paraphraser misunderstands the nuanced semantics and causes semantic drift that invalidates the sample's label.

%% file: 5_2_0_multiple.tex
\section{Multi-instance Level Data Augmentation}

Multi-instance augmentation combines value and structure information from different samples. 
The specific operation varies depending on the data modality in which it is applied. 

Multi-instance level image augmentation combines the visual information, i.e., pixels, from different images. It either mixes the colors of images or puts their patches together.
Multi-instance level text augmentation integrates the linguistic information from different sentences. It either mixes the sentence at the word level or combines fragments from multiple sentences. 
Multi-instance level graph augmentation mixes the network information of different graph components or graph instances. It either mixes the attributes of different nodes or combines both the topology and attributes of graphs.
Multi-instance level tabular augmentation combines the information of different lines or tables. Most methods mix the value representations across different lines, while it is also possible to combine two heterogeneous tables. 
Multi-instance time-series augmentation mixes values from different sequences, with few methods combining sequences using structural relationships. 

\vspace{-2mm}

\subsection{Value-based Mixture}

\input{5_2_1_multiple_value}

\vspace{-2mm}
\subsection{Structure-based Combination}
% \vspace{-2mm}

\input{5_2_2_multiple_structure}

\vspace{-4mm}
\subsection{Value-structure Mixture}

\input{5_2_3_multiple_val_struct}

%% file: 5_2_1_multiple_value.tex
\subsubsection{\textbf{Image Mixup}}

Mixup~\cite{zhang2017mixup} mixes two random images on the input layer by calculating their interpolation pixel-wise. Pixels from two original images each contribute a portion to the new image: 
\(
\mathbf{\Tilde{x}} = \lambda\mathbf{x}_i + (1-\lambda)\mathbf{x}_j, 
\mathbf{\Tilde{y}} = \lambda\mathbf{y}_i + (1-\lambda)\mathbf{y}_j
\),
where $(\mathbf{x}_i, \mathbf{y}_i)$ and $(\mathbf{x}_j, \mathbf{y}_j)$ are two labeled images from the dataset, and $\lambda \in [0,1]$.
Manifold mixup~\cite{verma2019manifold} is operated on the hidden layers of the neural network. 
AugMix~\cite{hendrycks2019augmix} applies three operations to one image and combines the three augmented images to generate a new sample. 
SmoothMix~\cite{lee2020smoothmix} applies a soft-edged mask to mix different regions in varied strength. 
Most mixup methods create new samples with soft labels, while sample pairing~\cite{inoue2018data} uses the label of one of the original images, and that of the other image is left unused. 
Similarly, ReMix~\cite{chou2020remix} addresses the class imbalance problem by assigning the minority class label to augmented data. 

\subsubsection{\textbf{Text Mixup}}

Unlike image data, whose coloring information is expressed by numerical values and can be interpolated at the input layer, words are difficult to interpolate directly, so text interpolation is usually done at the embedding layer. 
The adaptation of Mixup in sentence classification is studied in \cite{guo2019augmenting}, which proposes wordMixup and senMixup. 
Its wordMixup interpolates a pair of word embeddings and then feeds the interpolation to the sentence encoder. 
On the other hand, senMixup interpolates a pair of sentence embeddings. 
The use of mixup in text data augmentation has also been studied in other research, such as nonlinear Mixup \cite{guo2020nonlinear} and DoubleMix \cite{chen2022doublemix}. 

\subsubsection{\textbf{Graph Propagation}}

Propagation refers to spreading information from one node to other nodes along the graph. 
These methods include feature propagation and label propagation. 
Feature propagation shares node features with other nodes. 
After feature vectors of some nodes are masked out, GRAND~\cite{feng2020graph} further augments the node features by randomly propagating them through edges. 
MV-GCN~\cite{yuan2021semi} first creates three different views from the original graph, propagates the features in each, and then combines the three representations. 
The label propagation algorithm (LPA)~\cite{zhu2002learning} propagates labels from labeled nodes to all unlabeled nodes according to their proximity through areas with dense unlabeled nodes. 
AutoGRL~\cite{sun2021automated} applies LPA in its framework to label more nodes utilizing a few ground truth nodes. 
\cite{huang2020combining} performs two types of label propagation, correcting the base predictions and smoothing the final prediction, respectively. 

\subsubsection{\textbf{Tabular Mixup}}

Due to its well-aligned structure, tabular data is very suitable for interpolation.
SMOTE~\cite{chawla2002smote} is among the first works to study data interpolation. It addresses the problem of class imbalance by oversampling from the minority class. The samples are selected from the $k$ nearest neighbors in the feature space, and new synthetic data are produced by a random interpolation operation. 
As argued in \cite{darabi2021synthesising}, SMOTE is subject to unrealistic samples when interpolating in a sparse feature space. 
To solve this problem, TAEI~\cite{darabi2021synthesising} first encodes the sparse data into a dense latent space, interpolates the new data, and then decodes them to obtain the original data plus the augmented data. 
Contrastive Mixup~\cite{darabi2021contrastive} applies mixup for tabular data in a self- and semi-supervised learning framework. 
SAINT~\cite{somepalli2021saint} combines CutMix~\cite{yun2019cutmix} with contrastive learning to learn from tabular data. 
ExcelFormer~\cite{chen2023excelformer} proposes two schemes, namely Feat-Mix and Hidden-Mix. They interpolate data in the input layer and the hidden layer, respectively.

\subsubsection{\textbf{Sequence Mixing}}

DTW Barycenter Averaging (DBA)~\cite{petitjean2011global} revolves around the averaging problem for Dynamic Time Warping. It develops a global technique for computing the average of a set of sequences and avoids using iterative pairwise averaging.
In Weighted-DBA~\cite{forestier2017generating}, instead of each time series contributing equally to the final average, some can contribute more than others, so that an infinite number of new examples from any set of given time series can be generated. 
\cite{devries2017dataset} applies both interpolation and extrapolation in the feature space to augment time-series data. It first uses a sequence autoencoder to learn a feature space from unlabeled data. The data is then encoded and augmented with additive noise, interpolation, and extrapolation. Finally, the data in feature spaces is decoded and classified. 
Extending sequence mixing to the frequency domain, \cite{nanni2020data} applies the Equalized Mixture Data Augmentation (EMDA) to create new data. It computes the weighted average of two randomly chosen spectrograms with the same label.

%% file: 5_2_2_multiple_structure.tex
\subsubsection{\textbf{Image patching}}

Image patching puts together small fragments of many images or puts a part of an image over a second image. 
RICAP~\cite{takahashi2019data} randomly crops four images and patches them to construct a new image. 
CutMix~\cite{yun2019cutmix} is developed based on Cutout~\cite{devries2017improved}, but the removed region is replaced by a patch from another image. 
Attentive CutMix~\cite{walawalkar2020attentive} further enhances the CutMix method by adding a feature extraction module that selects an important or representative region and pastes this attentive patch onto the other image. 
CowMask~\cite{french2020milking} introduces patching with an irregular shape rather than a rectangular one. 

\subsubsection{\textbf{Text Fragment Merging}}

Text fragment merging combines text data by merging text fragments from one sentence into another. 
GECA~\cite{andreas2019good} removes a fragment from a sentence and populates it with another fragment that occurs in some common environment. 
Subtree Swapping~\cite{dehouck2020data}, Substructure Substitution~\cite{shi2021substructure}, and TreeMix~\cite{zhang2022treemix} use tree structures to analyze the part of speech or role in the sentences and thus make reasonable swaps. 
SSMix~\cite{yoon2021ssmix} replaces a low-saliency segment in a sentence with a segment with high saliency from another sentence and mixes sentence labels according to the ratio of their span length in the combined sentence. 

\subsubsection{\textbf{Table Concatenation}}

Contrary to single-instance level tabular augmentation, which lacks structural information to use, multi-instance level tabular augmentation suffers from strict and fixed table structures. Columns differ in their nomenclatures and representations, so it is hard to combine and learn from multiple distinct tables. If these restrictions are relaxed, models can learn across different tables, be more transferable, and better predict unseen tables. 
TransTab~\cite{wang2022transtab} suggests that multiple tables could share partially overlapped columns in the real world, but traditional methods often fail to fully utilize the data due to the removal of non-overlapping columns and mismatched samples when cleaning the data. TransTab first divides the table into three parts, each containing categorical, binary, and numerical data. These three types of data undergo different operations, are all converted to embeddings, and are again integrated into a new and encoded table. The processed tables can be joined in many ways, depending on the application scenarios. The augmented table contains information from many original tables and helps the model learn across them.

%% file: 5_2_3_multiple_val_struct.tex
\subsubsection{Image Value-structure Mixture}
\vspace{-2mm}

CNN models merge images by fusing pixels through channels and across the region within the reception fields. 
Diffusion models take textual prompts as inputs and can perform a flexible and smart mixture of existing images. 

\paragraph{\textbf{Neural Blending}}
AutoMix~\cite{liu2022automix} integrates a mix block in the training process. It builds a bridge between the selection of the mixup policy and the optimization of the model. 
Another image blending technique is proposed as Smart Augmentation~\cite{lemley2017smart}. 
At its input end, two random samples are fed into the network through two channels, and a CNN sub-network yields a one-channel output with the same shape as the input, which means one new image is generated based on the two input images. 
The augmented image is then used for downstream tasks, and the augmentation sub-network adjusts to the task performance accordingly.

\paragraph{\textbf{Prompt-based Mixing}}
Conditioned on textual inputs, diffusion models can generate combined images by using textual descriptions of multiple images. Diff-Mix~\cite{wang2024enhance} uses text embeddings of different categories to replace the foreground to align with the target classes. Diff-II~\cite{wang2024improving} calculates the inversion noise of two images, conducts random circle interpolation, and then denoises with different prompts. 

\subsubsection{Text Value-structure Mixture by \textbf{Generative Mixing}}
Generative mixing leverages the text understanding and generation capabilities of LLMs to combine information from multiple sentences. 
GPT3Mix~\cite{yoo2021gpt3mix} makes use of the latest large language model to mix sentences. 
By inserting two source sentences in a prompt template that describes the instruction to the model, we can get a mixed sentence of high quality with a soft label. 
PromptMix~\cite{sahu2023promptmix} also prompts the LLM to combine information from two sentences and then uses another LLM to relabel the generated augmentation sentence. 
Combining information from two text pieces is a subtle problem, and the specific behaviors of the LLMs highly depend on the prompts, which include the instruction, task description, and few-shot examples. So, the users have to try hard to design an ideal prompt so that the model works well.

\subsubsection{Graph Value-structure Mixture by \textbf{Graph Mixup}}
\label{sec:graph-valstr}

Graph mixup refers to mixing graph nodes or graph instances. 
This often involves both node attribute values and the topology structures. 
For node-level mixup, a topology-involved method~\cite{wang2021mixup}, randomly pairs nodes and includes the nodes' local topology representations when interpolating them. 
Structural Mixup~\cite{kim2023s} considers neighboring edges when pairing nodes and discusses how to connect the node with the original graph with edges. 
For graph-level mixup, ifMixup~\cite{guo2021ifmixup} adds dummy nodes so that two graphs have the same number of nodes. 
Graph Mixup with Soft Alignments~\cite{ling2023graph} obtains an assignment matrix by a soft assignment and uses that matrix to transform a graph so that every node corresponds with a node in the other graph. 
G-Mixup~\cite{han2022g} first uses graphs within the same class to estimate a graphon, i.e., graph generator: \(
    \mathbf{\mathcal{G}} \rightarrow W_\mathbf{\mathcal{G}},
    \mathbf{\mathcal{H}} \rightarrow W_\mathbf{\mathcal{H}}
\).
The graphon describes the probability of an edge between nodes $i$ and $j$. 
Graphons of two classes of graphs are then interpolated to produce a mixed graphon: \(
    W_\mathbf{\mathcal{I}} = 
    \lambda W_\mathbf{\mathcal{G}} + 
    (1-\lambda) W_\mathbf{\mathcal{H}}
\).
By sampling from the graphon mixup $W_\mathbf{\mathcal{I}}$, we acquire graphs with the same soft labels: \(
    \mathbf{y}_\mathbf{\mathcal{I}} = 
    \lambda \mathbf{y}_\mathbf{\mathcal{G}} + 
    (1-\lambda) \mathbf{y}_\mathbf{\mathcal{H}}
\).
DAGAD~\cite{liu2022dagad} mixes graph representations by performing random permutations, demonstrating that data augmentation is helpful in scenarios such as graph anomaly detection.

%% file: 5_3_0_populational.tex
\vspace{-3mm}
\section{Dataset Level Data Augmentation}
\label{sec:dataset}

Dataset-level data augmentation methods learn the overall data distribution and often generate new data that is consistent but diverse in terms of both value and structure information. 

\vspace{-3mm}
\subsection{Vanilla Generation}

A straightforward way to produce augmented data from the entire dataset is to learn the data distribution of existing datasets with neural networks and generate new samples that fall in that distribution. Some typical solutions are autoencoders (AE), generative adversarial networks (GAN), large language models (LLMs), and generative diffusion models. 

\input{5_3_1_vanilla}

\vspace{-4mm}
\subsection{Exogenous Generation}

Apart from the source dataset, we can also refer to exogenous information for data augmentation. Exogenous information is perceived by introducing external datasets or using knowledge learned by human experts.

\input{5_3_2_exogenous}

%% file: 5_3_1_vanilla.tex
\subsubsection{Autoencoder}

An autoencoder learns from the dataset and extracts the essential structure, which can be used to generate a more robust dataset. It is widely used for tabular data augmentation. 
\cite{delgado2021deep} tries out four types of autoencoder, concluding that all of these autoencoders can generate data sets that produce better task performance, while the variational autoencoder (VAE) provides the most robust result. Some other works \cite{fang2022semi,tran2019bayesian} also use VAE as their data generation approach. SDAT \cite{fang2022semi} adds extra noise to the latent space of its VAE to obtain augmented samples. These samples, along with the original ones, are then trained under a semi-supervised framework. \cite{tran2019bayesian} combines VAE with active learning. The samples are iteratively selected by active learning, labeled by an oracle, reconstructed by the VAE, and discriminated and classified. 
The use of AEs for feature space augmentation on image, text, and time-series data is discussed in \cite{devries2017dataset}, \cite{shorten2019survey}, and \cite{stocksieker2024data}. However, few follow-ups use such methods, potentially due to the inferior capacity compared with more advanced generation models. For instance, VAEs fail on complex datasets primarily because the limited approximation capacity of the encoder results in a poor, often overly simplistic, posterior distribution~\cite{bond2021deep}. When applied to graph data, VAE models also require a massive graph-matching process~\cite{liu2023generative}.

\subsubsection{Generative Adversarial Networks}

GANs enable generating new images on a large scale, leading to a new path of data augmentation. 

For image data, GANs have now been a common data augmentation technique, especially in data-scarce scenarios like the medical field~\cite{frid2018gan,kaur2021mr}. Apart from simply enlarging the dataset, GAN can be used in other ways, such as removing background noise and cleaning the dataset \cite{kupyn2018deblurgan,sharma2019learning,ashraf2021underwater}. It can also generate different views or poses of the object based on the original image \cite{zhang2022person}, and thus add more variation to the dataset. 
Despite the huge success of GAN's application for image data augmentation, a benchmark on biomedical image augmentation indicates that GAN-based methods do not generally outperform conventional data augmentation, very likely due to their limited generalizability~\cite{mantegna2024benchmarking}. 

Tabular data augmentation is another case where GANs play an important role. Table-GAN \cite{park2018data} mainly considers the privacy issue, and experiments show that the augmented table given by GAN is compatible with downstream machine learning models and produces similar performance. CTGAN \cite{xu2019modeling} focuses on the class imbalance problem and proposes a conditional generator that accounts for the imbalance in categorical columns. \cite{engelmann2021conditional} shares a similar goal and method with CTGAN \cite{xu2019modeling} but employs a more sophisticated loss function that better incorporates the discriminator and the auxiliary classifier. ITS-GAN \cite{chen2019faketables} proposes a framework that aims to preserve functional dependencies in tables. It models functional dependencies with autoencoders and includes a term to characterize the set of functional dependency constraints in the generator loss function. 
Apart from application to general tabular data, \cite{lacan2023gan} extends the use of GAN to transcriptomics data, where each row of the table represents a sample, each column represents a gene, and the table's values record the genes' expression levels. 

GANs are also widely used to generate virtual time-series data. \cite{esteban2017real} uses an RGAN and an RCGAN to produce realistic medical time-series data. TimeGAN \cite{yoon2019time} combines supervised and unsupervised losses to train its autoencoding components and adversarial components. \cite{donahue2018adversarial} propose WaveGAN and SpecGAN to generate time-series data. WaveGAN is used on the waveform, while SpecGAN is used on the spectrogram. SpecGAN demonstrates how GAN is applied to generate time-series data in the frequency domain. 

Although it is possible to generate graph data with GANs, \cite{de2018molgan} points out that GANs on graphs easily fall into mode collapse, where the generators tend to output non-unique samples, limiting GANs' application on graph datasets.

\subsubsection{Large Language Models}

Recent advances in large language models have established a new paradigm of data augmentation, especially benefiting the NLP field. 

LAMBADA~\cite{anaby2020not} is among the first attempts to use a state-of-the-art language generator to synthesize new text data for data augmentation. It generates a large dataset using a set of labels. 
Similarly, \cite{quteineh2020textual} uses the language model to generate a tree of tokens as candidates and constructs a sentence from them, and repeatedly refines the new sentences to build the augmented dataset. 
Data Boost~\cite{liu2020data} proposes a reinforcement learning framework to guide the generation process of GPT-2 to generate high-quality and high-similarity data. 
The rise of GPT-3 and later LLMs is now providing more possibilities for text generative augmentation, such as \cite{balkus2022improving} and \cite{cochran2023improving}. 

The development of multi-modal LLMs also facilitates image data augmentation. 
DIAGen~\cite{lingenberg2024diagen} uses the GPT-4 model to provide image descriptions with additional meaningful context and uses the enriched textual descriptions to generate diverse image datasets.
\cite{SAPKOTA2024100614} designs a variety of textual prompts and utilizes the DALL-E model to produce an exclusively LLM-generated agricultural dataset, avoiding traditional fieldwork with imaging sensors. 

LLMs can also be applied to tabular data augmentation by treating the table as text content. 
CLLM~\cite{seedat2024curated} uses the background description, in-context examples, and instruction guidelines to construct prompts and use the LLM to generate a large synthetic dataset, followed by a curator that evaluates and selects samples by their learning dynamics. 
Pred-LLM~\cite{nguyen2024generating} first converts the table rows to text descriptions to fine-tune an LLM. Then it masks some features of sampled rows and uses the LLM to recover these rows. Finally, it uses the LLM to label the generated data. 

Time-LLM~\cite{jin2024time} demonstrates the predictive ability of Transformer-based LLMs for time-series data, suggesting another possibility of LLMs as a data augmentation technique. However, beyond simply using LLMs to predict and expand sequences to future timestamps, the use of LLMs for general time-series data augmentation is still underexplored. Some exceptions are in the audio, voice, and speech fields, where the language information can be utilized by LLMs, enabling more possible manipulations, such as ArzEn-LLM~\cite{heakl2024arzen}, which employs the code-switched augmentations for multilingual speech recognition performance. 

\subsubsection{Generative Diffusion Models}

Diffusion models have recently emerged as a new solution to generative AI. Generating realistic and diverse samples by learning the underlying data distributions, they also show promising capabilities as a tool for data augmentation.

The diffusion model is widely applied to image data. 
GIF~\cite{zhang2023expanding} leverages Stable Diffusion to optimize the latent features of the seed data in the semantically meaningful space of the prior model and create informative and realistic new data. 
SIGD~\cite{li2024semantic} employs diffusion models to generate augmented images with good image diversity. It takes image labels and captions as guidance to maintain semantic consistency. 
\cite{li2024simple}  prompts the diffusion models to ``generate a clean background'' to avoid additional objects in the background and designs a background mask erosion mechanism to avoid extending objects to the background. 

For text data, DiffusionCLS~\cite{chen2024effective} leverages a diffusion LM to capture in-domain knowledge and generate pseudo samples by reconstructing from the corrupted dataset to balance between consistency and diversity. 
DiffLM~\cite{zhou2024difflm} leverages diffusion models to reserve more information of the dataset's original distribution and format structure in the learned latent distribution and decouples the learning of target distribution knowledge from the LLM’s generative objectives via a plug-and-play latent feature injection module. 

The diffusion model also fills the gap of tabular data augmentation. 
\cite{liu2024controllable} leverages diffusion models to first learn an unconditional generative model and introduces lightweight controllers to guide the unconditional generative model in generating synthetic data that satisfies different conditions.
TabDiff~\cite{shi2024tabdiff} develops a joint continuous-time diffusion process for numerical and categorical data and proposes feature-wise learnable diffusion processes to counter the high disparity of different feature distributions. 
Forest-Diffusion~\cite{jolicoeur2024generating} introduces a novel approach for generating and imputing mixed-type tabular data utilizing score-based diffusion and conditional flow matching.

Compared with other modalities, using the diffusion model for graph data and time-series data augmentation is relatively underexplored. 
To implement diffusion models on graphs, some properties of graphs must be properly treated. For example, diffusion models often rely on a continuous Gaussian noise process, but the graph structures are discrete~\cite{zhang2023survey}. In response, DiGress~\cite{vignac2023digress} designs a discrete denoising diffusion model for graph generation and achieves state-of-the-art performance on molecular and non-molecular datasets. 
Future advancement of generative diffusion models for graph data augmentation will especially benefit studies on molecules, proteins, and materials, whose data are often in the form of graph instances. 
For tabular data, \cite{liu2024adaptive} proposes ASE-DDPM for unbalanced time series classification. 
\cite{zhang2025time} employs the diffusion model for time-series data but only uses it to generate the first moment data. 
Despite these attempts, literature focusing on the design and application of diffusion-based time-series data augmentation is still scarce. 
Given the remarkable capability of diffusion models and their success in other data modalities, it is promising to further delve into this field.

%% file: 5_3_2_exogenous.tex
\subsubsection{Introducing Exogenous Datasets}

\paragraph{\textbf{Neural Style Transfer}}
Image datasets vary in artistic styles, and neural style transfer~\cite{gatys2016image} introduces style information from an external dataset when augmenting an entire dataset. It learns the artistic style from a group of image data and uses this artistic style to depict the topic of images from another dataset. As is \cite{gatys2016image}, neural style transfer can be used to mimic the drawing styles of particular artists. Such an augmentation process can help the model learn a more general representation of the topics. \cite{li2018closed} proposes a more realistic use of neural style transfer. It transfers the weather or illumination from reference images to the target dataset. This is particularly useful because it is hard to collect sufficient images under certain weather.

\paragraph{\textbf{Table Retrieval}}
Tabular data populates the web, and by retrieving consistent and related tables, it is possible to generate an augmented table that incorporates extended information. 
A key consideration of table retrieval augmentation is to design appropriate ways to search for ideal tables. 
Santos~\cite{khatiwada2023santos} leverages external knowledge bases, YAGO 4, and focuses on designing a union search technique to match desired target tables. 
HYTREL~\cite{chen2023hytrel} predicts table similarities by building hypergraphs on top of the tables and uses a hypergraph encoder to calculate the encoding. 
LEKA~\cite{zhang2025leka} uses an LLM to extract the table's descriptive information, retrieve it in the Kaggle database, and harmonize tables from different sources.

\subsubsection{Leveraging Expert Knowledge}

Here we introduce three examples of how humans utilize expert knowledge for image, tabular, and time-series data augmentation

\textit{\textbf{Computer Graphic Modeling}} is to leverage expert knowledge on how objects in the existing dataset look and the power of computer software to generate new image samples \cite{gaidon2016virtual,mccormac2017scenenet}. These datasets have several advantages over ordinary image datasets. First, it requires little human effort to collect data. Second, by applying different aspects, poses, and lighting conditions, computer graphics modeling can generate images with high diversity. Third, these images are all fully labeled with accurate ground truth. 

\textit{\textbf{Relational Structure Construction}} can enhance the potential of tabular data, which is limited by its simple nature compared with data from other modalities. Due to the lack of spatial or relational relationships, many popular augmentation approaches and network configurations are not suitable for tabular datasets. A solution is to construct and introduce a relational structure based on the tabular dataset.
GOGGLE \cite{liu2022goggle} proposes a solution to this problem. It learns the relational structure and corresponding functional relationships from tabular datasets and uses this basis to generate new samples. In other words, it learns a graph from the features in the table and obtains augmented data from that graph. PET \cite{du2022learning} depicts multiple tabular data instances as a hypergraph and uses the labels to help construct hyperedges. These methods are based on external assumptions and introduce extra structural information beyond the tabular dataset. 

\textit{\textbf{Statistical Generation}} is a technique that predicts values in the future. 
Some methods view time-series data outside the existing data and from a statistical perspective. They are based on human knowledge of statistics and can predict how the value changes in future timestamps. 
\cite{smyl2016data} uses a statistical algorithm called LGT (Local and Global Trend) to forecast paths of the time-series data and samples from them to get extra data. 
GRATIS \cite{kang2020gratis} employs mixture auto-regressive (MAR) models to generate collections of time series and examine the variety and extent of the produced time series within a feature space for time series.

%% file: 9_discussion.tex
\vspace{-3mm}
\section{Discussions}
\vspace{-3mm}

Up to now, this survey focuses on wherefrom the information in the augmented samples is derived. And there are many other perspectives to discuss data augmentation and understand how it works. 

\vspace{-4mm}
\subsection{Augmentation Evaluation Metrics}
\vspace{-2mm}

Most research on data augmentation uses the models' performance on downstream tasks to evaluate the effect of data augmentation~\cite{cubuk2019autoaugment}. 
However, this criterion depends on many extraneous factors and may fail to reflect the influence of data augmentation methods on the data. 
So, there is a need for a set of evaluation metrics that can assess specific considerations of data augmentation. 
Two of the most critical metrics are affinity and diversity~\cite{alihosseini2019jointly,gontijo2020tradeoffs,yang2024investigating}. 

Affinity is also referred to as consistency, correctness, faithfulness, quality, similarity, or validity. It reflects if the augmented sample is semantically consistent with the original sample and preserves the original label. 
Model-centric evaluation~\cite{gontijo2020tradeoffs} defines affinity as the ratio of model \(m\)'s performance on the augmented validation dataset \(D^{\prime}_{val}\) and that on the original validation dataset \(D_{val}\), given by
$\mathit{affinity} = \text{ACC} (m, D^{\prime}_{val}) / \text{ACC} (m, D_{val})$,
where \(\text{ACC} (m, D)\) stands for model \(m\)'s performance on the evaluation dataset \(D\). 
Data-centric evaluation, on the other hand, defines affinity as how much the augmented dataset deviates from the original dataset in the feature space. For example, \cite{alihosseini2019jointly} proposes Fr\'echet BERT Distance (FBD) to evaluate the deviation of two textual datasets, defined as: 
$\mathit{FBD} = \sqrt{\|m_1 - m_2\|^2_2 + Tr(C_1 + C_2 - 2(C_1C_2)^{1/2}}$,
where \(m\) is the mean vector and \(C\) is the covariance matrix. 

Diversity reflects the breadth of patterns in the dataset, and exposing the model to a diverse training dataset can improve its robustness. 
\cite{lai2020diversity} considers a dataset as an ellipsoid and defines its diversity as the isocontour radius of the ellipsoid: 
$\mathit{diversity} = (r_1{\cdot}r_2{\cdot}...{\cdot}r_H)^\frac{1}{H}  = \sqrt[H]{\prod^H_i \sigma_i}$,
where \(H\) is the dimension of the feature vectors and \(\sigma_i\) is the standard deviation along the \(i\)-th feature. 
\cite{yu2022can} proposes to evaluate dataset diversity by calculating the dispersion:
$\mathit{disersion} = \sum_{x_i,x_j} d(x_i,x_j)$, 
where \(d(x_i,x_j) = 1 - \frac{f(x_i){\cdot}f(x_j)}{\|f(x_i)\|\|f(x_j)\|}\). 

Some empirical studies investigate how affinity and diversity influence the effectiveness of data augmentation, and their real-world experimental results demonstrate that model performance takes the most advantage of data augmentation that achieves both high affinity and diversity~\cite{gontijo2020tradeoffs,yang2024investigating,yu2022can}. 

\vspace{-4mm}
\subsection{Selecting the Target}
\vspace{-1mm}

\subsubsection{On Selecting Pixels}
\label{sec:sel_pix}

For pixel erasing, selecting target pixels is a necessity before erasing them. Image-aware Random Erasing (IRE)~\cite{zhong2020random} randomly selects an area in the image. 
Object-aware random erasing (ORE)~\cite{zhong2020random} restricts random erasing to the object's bounding box if its location is known.
\cite{yang2019region} proposed Region-aware Random Erasing (RRE), which introduces ORE with upper bounds for width and height in the bounding boxes, and then randomly erases the background.
The same consideration also applies to image cropping (IC) and mixup, and some importance-based strategies are used to preserve important contents in the images. 
For IC, \cite{ma2004automatic} uses entropy as the importance score to select the most important regions, and \cite{santella2006gaze} uses eye-tracking technique to locate important content. 
For mixup, Co-mixup~\cite{kim2021co} and Puzzle Mix~\cite{kim2020puzzle} both use saliency to decide the mixing strength of different regions. 
A summary of how these methods select target pixels is presented in Table~\ref{tab:img_region_selection}.
\vspace{-2mm}

\input{tab/img_region_selection}

\vspace{-2mm}
\subsubsection{On Selecting Words}

Most value-based text transformation methods randomly select the word to be replaced, added, and deleted, but some methods consider the role a word plays in the sentence when deciding whether to impose perturbation.

For example, TEXTFOOLER~\cite{jin2020bert} ranks words by their importance and replaces the important words with their synonyms.
Syntax-aware data augmentation~\cite{duan2023syntax} improves upon older methods by modifying words based on their grammatical role in a sentence. It uses a sentence's structure (a dependency tree) to determine a word's importance; words less central to the sentence's core meaning are more likely to be changed or removed.
Selective Text Augmentation (STA)~\cite{guo2022selective} ignores grammatical structure and instead classifies words based on their statistical and semantic relationship to a category. It then groups words into four roles (like "gold" for important words). During augmentation, it treats words differently based on their role; for example, it will keep "gold" words but add synonyms for them.
A summary of how these methods select target words is presented in Table~\ref{tab:text_word_selection}.
\vspace{-2mm}

\input{tab/text_word_selection}

\subsubsection{On Selecting Nodes and Edges}

For graph data augmentation, a key problem is selecting target nodes and edges to operate on. 
Node centrality acts as an important factor that is both meaningful and convenient, supporting selection strategies of many graph data augmentation methods. 
GCA~\cite{zhu2021graph} uses node centrality to calculate the masking vectors and rewiring probabilities and thus decides which node to mask and which edge to rewire. 
GraphCrop~\cite{wang2020graphcrop} applies a node-centric strategy to preserve connectivity in subgraphs.
SUBG-CON~\cite{jiao2020sub} uses the Personalized PageRank centrality to select the important nodes for subgraphing. 
Smart rewiring~\cite{louzada2013smart} takes the degrees of chosen nodes when swapping their edges. 
Another useful factor is node similarity. 
DHGR~\cite{bi2022make} learns a similarity matrix from the original graph and rewires the graph by adding edges between high-similarity node pairs and removing edges from low-similarity ones. 
Similarly, RobustECD~\cite{zhou2021robustecd} and M-Evolve~\cite{zhou2020data} also consider node similarity for edge rewiring. 

\input{tab/graph_node_selection}

\vspace{-4mm}
\subsection{Leveraging Augmented Samples}

A basic use of data augmentation is to enlarge the dataset, especially in data-scarce scenarios, where the augmented samples are appended to the training dataset. 
Beyond this paradigm, data augmentation provides many other possibilities to support deep learning. 
A frequent usage of augmented samples is to construct different views for contrastive learning.
Rock-ViT~\cite{wang2024deep} performs rotation, flipping, blurring, contrast enhancement, irregular noise, and microscope dark-field noise on sandstone microscopic images. It optimizes the vision transformer with both cross-entropy loss and contrastive loss. When calculating the contrastive loss, Rock-ViT uses two random augmentations to build positive pairs. 
Instead of simply calculating contrastive loss between positive and negative sample pairs, DualCL~\cite{chen2022dual} augments text samples by including the label in the sentence. Specifically, for an original sentence ``[CLS] love this movie" in a sentiment classification dataset, DualCL converts it to ``[CLS] POS NEG love this movie", where ``POS" and ``NEG" denote the positive and negative labels. Then it repels the embeddings of the ``[CLS]" token and the ``NEG" token, since this sample is possible and thus ``[CLS] should also be positive. Conversely, it attracts the embeddings of the ``[CLS]" and ``POS" tokens. 
Contrastive learning is a common practice in graph deep learning, and graph data augmentation plays a critical role in many graph deep learning pipelines. 
GraphCL~\cite{you2020graph} uses Node Dropping and Edge Perturbation to build different views, and MVGRL~\cite{hassani2020contrastive} applies graph diffusion. Different views of an original graph are used as positive pairs, and a contrastive loss function is used to enforce maximizing their consistency. 
VIME~\cite{yoon2020vime} extends the usage of data augmentation beyond contrastive learning. It incorporates tabular data augmentation in both self-supervised and semi-supervised learning frameworks. For self-supervised learning, it corrupts the table with a mask and optimizes the model with table reconstruction loss and mask estimation loss. For semi-supervised learning, it calculates the consistency loss between masked views.

\vspace{-4mm}
\subsection{How to Choose the Appropriate Augmentation Method?}
\vspace{-1mm}
As summarized in many surveys and benchmarks on data augmentation, although different data augmentation methods claim to be effective and boast certain advantages, their practical performance largely depends on specific task scenarios and even nuanced experimental settings. So, it is hard to say which method is the best one. 
Still, we summarize some characteristics, either as advantages or limitations, of each method in Table~\ref{tab:summary}. Specifically, we highlight the methods that (1) perform sample-wise analysis (e.g., saliency map or syntax parsing) before augmentation, (2) require dedicated computation to generate the augmentation output, and (3) are subject to indistinctive or non-recoverable information loss during the augmentation processes. 
We also address some practical considerations when choosing augmentation methods under different circumstances in detail in this section. 
\subsubsection{Tradeoff between the Gain and Cost}
Many advanced data augmentation methods claim better performance compared with simple and random methods. However, these advanced methods often involve complicated techniques, such as reinforcement learning and generative models, which require even more time and resources to augment the dataset than training on the datasets. To balance the performance and cost, a key issue is what we expect from data augmentation. Suppose the current dataset is already very large; perhaps we just want to introduce some corruption to improve model robustness, then cheaper methods like image flipping, synonym replacement, and noise addition would be good choices. In case of low-resource scenarios where data is scarce, it might be wise to resort to generative methods for their capabilities to introduce diversity and expand dataset coverage~\cite{mantegna2024benchmarking,wang2025diversity}. 
\subsubsection{Be Careful with the Labels}
A crucial consideration when choosing the data augmentation method is the characteristics of the dataset's labels. 
The labels in some datasets are subject to change under some operations. For example, image rotation is not safe for the MNIST dataset, which contains ``6'' and ``9''. Similarly, random token insertion might be label-preserving for the SST-2 sentiment classification dataset but is harmful for the CoLA grammar acceptability dataset. 
Multi-instance augmentation, such as promptMix~\cite{sahu2023promptmix}, is a powerful tool to refine the decision boundary between categories for a classification dataset. However, it is hard to apply to unlabeled datasets in semi-supervised settings, where single-instance augmentation plays a dominant role. 
\subsubsection{Compatibility with the Training Framework}
Multi-instance augmentation is also challenged by some model training frameworks. Some models only accept hard-labeled data, while mixup-based methods often produce soft labels when mixing samples. 
If data augmentation is performed as an intermediate step in the model's framework, both input-space and hidden-space augmentation are available, while if augmentation is used before training the model, it would be hard to apply hidden-space augmentation because the embeddings have not been produced. 

\vspace{-4mm}
\subsection{Transferring between modalities}
\vspace{-1mm}
We have proposed a cross-modality taxonomy that unifies data augmentation for five common modalities. 
From this cross-modality perspective, we notice that many data augmentation methods designed for different modalities, though dubbed with different names, share the same nature with respect to their operation on samples' information. 
For example, pixel erasing, token deletion, node attribute masking, table masking, and sequence jittering all apply masks and introduce noise to the samples, and image cropping, sentence cropping, subgraph cropping, table subsetting, and sequence window slicing all break the structure of a sample. 
Conversely, such consistency implies that data augmentation techniques well developed for one modality can be transferred to other under-explored modalities, inspiring novel advancements in this field. 
Policy-based augmentation image combines multiple basic operations for image data~\cite{cubuk2019autoaugment,lim2019fast}, and since abundant atomic operations are also available for text and graph data, such as token replacement or edge dropping, it could be a very good idea to design agents that search for optimal text or graph augmentation strategies. 
Image patching and text fragment merging combine pieces from multiple images or sentences together as a new sample~\cite{yun2019cutmix,dehouck2020data}. To transfer this idea to tabular data, we could think about if it is possible to combine half columns from a row with the remaining columns from another row. 
Advanced neural networks have enabled a series of augmentation approaches that can generate high-quality and high-fidelity image and text data~\cite{trabucco2024effective,wang2025diversity}. Though yet underexplored, we could expect more similar advancements for graph, tabular, and time-series data.

\vspace{-4mm}
\subsection{Harnessing the Latest Techniques}
\vspace{-1mm}
Early data augmentation methods rely on hand-crafted manipulations, which are constrained by effectiveness and efficiency. AutoAugment~\cite{cubuk2019autoaugment} begins to introduce reinforcement learning to find optimal augmentation policies, enabling automatic search for desired augmentation strategies. Generative models, such as LLMs and diffusion models, possess outstanding capacities to capture sample semantics and produce new samples with high flexibility and high quality~\cite{cai2023resolving,wang2024improving,dai2025auggpt}. 
However, most generative data augmentation methods are constrained to simple deployment in their by-design modalities, for example, modifying images with diffusion models or paraphrasing sentences with LLMs. 
In response, we suggest two concrete directions: 
(1) Designing informed data augmentation methods that take into account the sample distribution and supervision power of the datasets. For example, DoAug~\cite{wang2025diversity} bridges the gap between LLM generation and datasets' sample diversity.
(2) Embracing the cross-modality nature of many generative models and utilizing the advantage of their representative and generative power in different modalities. For example, DALDA~\cite{jung2024dalda} uses the GPT-4 model to generate image captions and consequently employs the diffusion model to generate corresponding images. 

%% file: tab/img_region_selection.tex
\begin{table}[htbp]
    \centering
    \tiny
    \vspace{-3mm}
    \caption{Region selection strategy for pixel erasing, image cropping, and mixing methods.}
    \vspace{-3mm}
    \setlength{\tabcolsep}{4pt} % horizontal, Default value: 6pt
    \renewcommand{\arraystretch}{0}
    \begin{tabular}{p{3cm}p{2cm}p{3cm}}
\toprule
Random & Location-based & Importance-based \\
\midrule
Cutout~\cite{devries2017improved}, Image-aware RE~\cite{zhong2020random}, Region-aware RE~\cite{yang2019region}, Hide-and-seek~\cite{kumar2017hide}, Grid-Mask~\cite{chen2020gridmask} & Object-aware RE~\cite{zhong2020random}, Region-aware RE~\cite{yang2019region} & Entropy-bases IC~\cite{ma2004automatic}, Gaze-based IC~~\cite{santella2006gaze}, Co-mixup~\cite{kim2021co}, Puzzle mix~\cite{kim2020puzzle} \\
\bottomrule
    \end{tabular}
\label{tab:img_region_selection}
\end{table}

%% file: tab/text_word_selection.tex
\begin{table}[htbp]
    \centering
    \tiny
\vspace{-4mm}
    \caption{Word selection strategy of text data selection methods.}
\vspace{-2mm}
    \setlength{\tabcolsep}{2pt} % horizontal, Default value: 6pt
    \renewcommand{\arraystretch}{0}
    
    \begin{tabular}{p{1.2cm}p{4.5cm}p{2.5cm}}
\toprule
Random & Role-based & Learned \\
\midrule
EDA~\cite{wei2019eda}, \cite{marivate2020improving}, AEDA~\cite{karimi2021aeda}, \cite{xie2017data} &
TEXTFOOLER~\cite{jin2020bert}, LeCA~\cite{chen2021lexical}, GBS~\cite{hokamp2017lexically}, STA~\cite{guo2022selective}, Substructure Substitution~\cite{shi2021substructure}, Subtree Swapping~\cite{dehouck2020data}, TreeMix~\cite{zhang2022treemix} &
LetterTran~\cite{liu2011insertion}, INSNET~\cite{lu2022insnet}, SSMix~\cite{yoon2021ssmix} \\
\bottomrule
    \end{tabular}
\vspace{-2mm}
\label{tab:text_word_selection}
\end{table}

%% file: tab/graph_node_selection.tex
\begin{table}[htbp]
    \centering
    \tiny
    \vspace{-3mm}
    \caption{Node and edge selection strategy for grpah data augmentation methods.}
    \vspace{-3mm}
    \setlength{\tabcolsep}{4pt} % horizontal, Default value: 6pt
    \renewcommand{\arraystretch}{0}
    \begin{tabular}{p{2cm}p{3cm}p{2.5cm}}
\toprule
Random & Centrality & Similarity \\
\midrule
Attribute masking~\cite{you2020graph}, GRAND~\cite{feng2020graph}, BGRL~\cite{thakoor2021large} & 
GCA~\cite{zhu2021graph}, GrpahCrop~\cite{wang2020graphcrop}, SUBG-CON~\cite{jiao2020sub}, Smart rewiring~\cite{louzada2013smart} & 
DHGR~\cite{bi2022make}, RobustECD~\cite{zhou2021robustecd}, M-Evolve~\cite{zhou2020data} \\
 \bottomrule
    \end{tabular}
\vspace{-4mm}
\label{tab:graph_node_selection}
\end{table}

%% file: 10_conclustion.tex
\vspace{-6mm}
\section{Conclusion}
\vspace{-1mm}
This survey presents a comprehensive summary of data augmentation techniques across five data modalities and proposes a modality-independent taxonomy from a data-centric perspective, which focuses on from where the augmented data is derived. It further enumerates related research papers on these data augmentation methods and annotates them with descriptive information in detail. Finally, it discusses several critical issues related to data augmentation that can help us better understand this technique. 
\vspace{-3mm}